\documentclass[times,review,10pt]{elsarticle}
\usepackage{amsmath,amssymb,amsfonts}
\usepackage{algorithmic}
\usepackage{array}
\usepackage[caption=false,font=normalsize,labelfont=sf,textfont=sf]{subfig}
\usepackage{textcomp}
\usepackage{stfloats}
\usepackage{url}
\usepackage{verbatim}
\usepackage{graphicx}
\usepackage{cite}
\usepackage[T1]{fontenc}
\usepackage[ruled,linesnumbered]{algorithm2e}
\usepackage[colorlinks,linkcolor=blue, citecolor=blue,]{hyperref}
\usepackage{float}
\usepackage{booktabs}
\usepackage{multirow}
\usepackage{tabularx}
\usepackage{bbding}
\usepackage[dvipsnames]{xcolor}
\usepackage{colortbl}
\usepackage{adjustbox}
\usepackage[dvipsnames]{xcolor}

\usepackage{amssymb}
\usepackage{url}
\newcommand{\etal}{\textit{et al}. }

\begin{document}

\begin{frontmatter}

\title{Neural Spatial-Temporal Tensor Representation for Infrared Small Target Detection}

\author[inst1,inst2]{Fengyi Wu}
\ead{wufengyi98@163.com}
\author[inst1,inst2]{Simin Liu}
\ead{liusimin\_0719@163.com}
\author[inst2,inst3]{Haoan Wang}
\ead{whaan712@gmail.com}
\author[inst1,inst2]{Bingjie Tao}
\ead{taobingjie@uestc.edu.cn}
\author[inst1,inst2]{Junhai Luo}
\ead{junhai\_luo@uestc.edu.cn}
\author[inst1,inst2]{Zhenming Peng\corref{corresponding}}
\ead{zmpeng@uestc.edu.cn}
\cortext[corresponding]{Corresponding author.}

\affiliation[inst1]{organization={School of Information and Communication Engineering},
            addressline={University of Electronic Science and Technology of China}, 
            city={Chengdu},
            postcode={611731}, 
            country={China}}
\affiliation[inst2]{organization={The Laboratory of Imaging Detection and Intelligent Perception},
            addressline={University of Electronic Science and Technology of China}, 
            city={Chengdu},
            postcode={611731}, 
            country={China}}

\affiliation[inst3]{organization={College of Glasgow},
            addressline={University of Electronic Science and Technology of China}, 
            city={Chengdu},
            postcode={611731}, 
            country={China}}

\begin{abstract}
Optimization-based approaches dominate infrared small target detection as they leverage infrared imagery's intrinsic low-rankness and sparsity. While effective for single-frame images, they struggle with dynamic changes in multi-frame scenarios as traditional spatial-temporal representations often fail to adapt.
To address these challenges, we introduce a Neural-represented Spatial-Temporal Tensor (NeurSTT) model. This framework employs nonlinear networks to enhance spatial-temporal feature correlations in background approximation, thereby supporting target detection in an unsupervised manner. Specifically, we employ neural layers to approximate sequential backgrounds within a low-rank informed deep scheme. A neural three-dimensional total variation is developed to refine background smoothness while reducing static target-like clusters in sequences. Traditional sparsity constraints are incorporated into the loss functions to preserve potential targets. By replacing complex solvers with a deep updating strategy, NeurSTT simplifies the optimization process in a domain-awareness way. Visual and numerical results across various datasets demonstrate that our method outperforms detection challenges. Notably, it has 16.6$\times$ fewer parameters and averaged 19.19\% higher in $IoU$ compared to the suboptimal method on $256 \times 256$ sequences.

\end{abstract}


\end{frontmatter}

\section{Introduction}
\label{sec1}
Infrared small target detection (ISTD) has gained significant attention due to the robustness of infrared imagery in all weather conditions \citep{kou2023survey}. However, the long detection range often makes targets appear dim and tiny, lacking shape features, and having low signal-to-clutter ratios \citep{zhao2022single}. Consequently, detecting infrared targets in various scenarios remains a prominent academic challenge \citep{zhang2024irsam}.

ISTD methods are generally divided into traditional model-based and deep learning-based approaches. Optimization-based methods have garnered more attention among the model-based category, due to their interpretability via robust principal component analysis (RPCA) models \citep{gao2013ipi, dai2017ript}. However, these methods often struggle with the limitations of traditional matrix or tensor operators, which fail to efficiently correlate the background or target with linear representations in complex scenarios \citep{wu2024rpcanet}. Deep learning-based (DL) ISTD methods seldom meet such difficulties as they utilize artificial neural networks to map original images to ground truth labels. However, these methods often lack model guidance and suffer from the black-box nature of deep structures. How to combine the merits of these two categories, therefore theoretically modeling the detection task in a deep manner, has been a growing interest.

In addition, with the increasing availability of sequence infrared datasets, ISTD methods are now categorized into single-frame and multi-frame approaches. Unlike single-frame methods which only use spatial information, multi-frame methods utilize richer spatial and temporal information for background/target separation. Some researchers have attempted to use visible light priors to capture motion features \citep{kwan2020optical}. However, these approaches often fail as they overlook the characteristics of infrared imagery. Recognizing this, there is a growing focus on leveraging the low-rankness and sparsity in infrared imagery by utilizing temporal information within the optimization domain, typically, by stacking sequences in a high-dimensional tensor \citep{liu2021asttv}.
However, tensor-based algorithms struggle to balance effectiveness with efficiency, often requiring extensive time for tensor or matrix calculations. In addressing these, deep structures \citep{chen2024sstnet} have been introduced for temporal detection tasks, yielding more satisfactory results with fewer manual adjustments. Despite their success, most of these methods focus on the "target" and seldom consider the background representation in infrared imagery, leading to potential false alarms or missed detections when the environment changes. Additionally, these fully supervised methods require extensive mask or box labeling and significant computational resources for training on video data.

To tackle these challenges, recent studies have combined the aforementioned optimization models, which balance attention on background and target, with flexible deep networks. Techniques like deep unfolding networks (DUNs) and deep plug-and-play (PnP) are popular \citep{liu2022deeptensor}. DUNs unfold optimization processes within an interpretable deep framework but require extensive data and are inefficient for long videos due to high computational demands. Deep PnP methods are used as pre-trained components in inverse problem image tasks, but their reliance on pre-trained models can lead to false eliminations in new scenarios \citep{liu2023wswtnn-pnp} due to mismatches between infrared imaging and other resources. Thus, while the above two approaches complete the incorporation of deep networks and optimization models, their demands for extra training data remain a significant issue.

\begin{figure}[t!]
\setlength{\abovecaptionskip}{-2pt}
\setlength{\belowcaptionskip}{-2pt}
\centering
\includegraphics[width=0.7\linewidth]{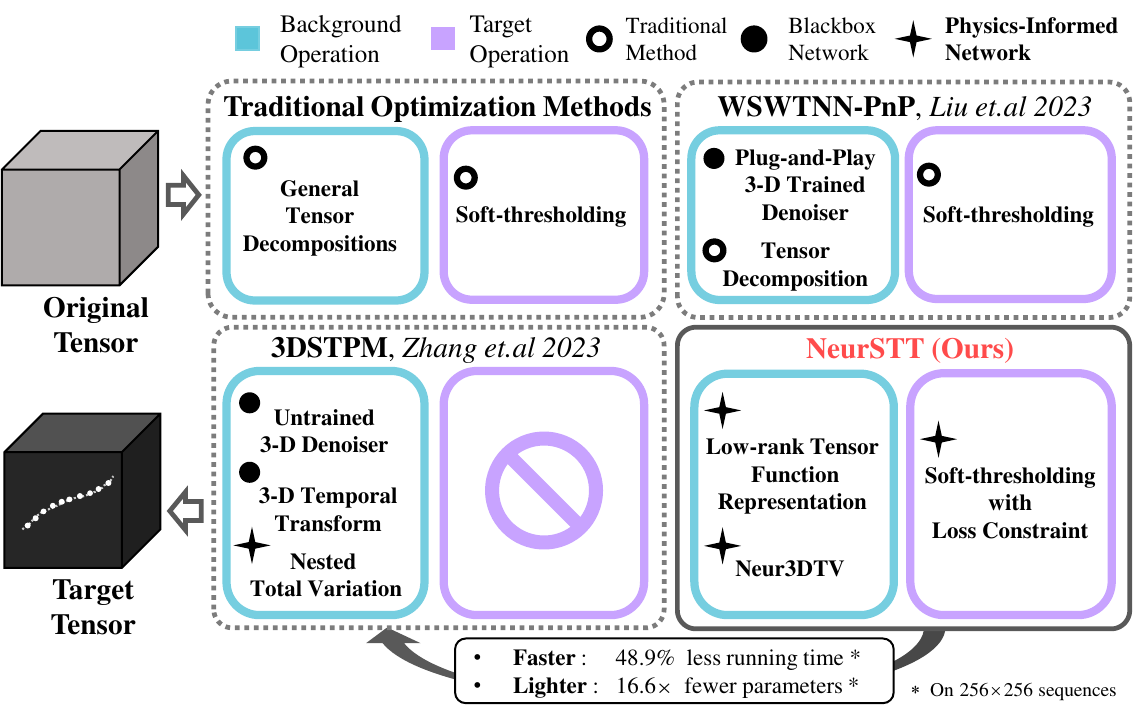}
\caption{Comparison of the existing spatial-temporal tensor schemes for ISTD and our NeurSTT with physic-informed (low rankness and sparsity) unsupervised learning strategy, while NeurSTT has fewer parameters and a faster execution time.}
\label{fig_hl}
\vspace{-0.1cm}
\end{figure}

Implicit neural representations (INRs) suggest that the network can iteratively represent a well-constructed image without additional labeled data and have proven effective in various vision tasks \citep{miao2022ds2dp}. However, many of these methods use basic neural constructions, such as three-dimensional (3-D) convolutional neural networks (CNNs), leading to high parameter consumption \citep{zhang20233dstpm} and gradient vanishing \citep{molaei2023implicit}. A representation network aware of data features, particularly from infrared imagery, is preferable. Moreover, unlike low-level vision tasks with static backgrounds, leveraging temporal information to separate backgrounds and targets is crucial for high-level detection. Thus, finding a proper neural representation to regularize these two key elements is essential. Benefiting from these less data-intensive deep strategies, we aim to develop a domain-specific approach to represent spatial-temporal features with proper regularization, assisting in the final detection task in ISTD.

In this article, we propose a neural-represented spatial-temporal tensor module (NeurSTT) for ISTD. Our approach nonlinearly approximates the background using a low-rank guided tensor network, reducing parameters by leveraging the inherent properties of infrared data, as shown in Fig. \ref{fig_hl}. Drawing inspiration from traditional ISTD methods, we employ a neural-based three-dimensional total variation (Neur3DTV) to capture local smoothness in both temporal and spatial domains, thereby enhancing target detection. Additionally, we update sparse targets using a soft-thresholding operator and incorporate it into the loss constraint. By combing different losses from three updating blocks, we solve NeurSTT using the adaptive moment estimation (Adam) algorithm \citep{kingma2014adam}. This model is capable of detecting sequential targets in an unsupervised learning manner, eliminating the need for labeled data.
Extensive evaluations of various datasets demonstrate the feasibility and effectiveness of our model, showing superior performance compared to baselines.

The contributions of this study to ISTD can be summarized as follows: 
\begin{enumerate} 
\item We propose a low-rank informed background approximation network to enhance spatial-temporal feature representation. The powerful nonlinearity of neural networks assists in this estimation, which we further constrain using a nuclear norm-based learning loss.
\item Instead of relying on discrete differential regularizations, we develop Neur3DTV, a strategy to capture spatial and temporal local correlations through continuous representation in deep neural networks (DNNs).
\item We incorporate the traditional soft-thresholding operator into the updating scheme via loss constraints, rather than directly subtracting the observed tensor from the background.
\item This study addresses NeurSTT using a deep updating strategy in an unsupervised manner, aggregating loss functions from corresponding modules. Numerical and visual results on various datasets demonstrate the algorithm's superior performance with compared baselines.
\end{enumerate}

In the following sections, we review related work on ISTD in Section \ref{sec2} and outline necessary notations and preliminaries in Section \ref{sec3}. Section \ref{sec4} details the solution and construction of the NeurSTT model. Quantitative results and ablation studies demonstrating the algorithm’s effectiveness are presented in Section \ref{sec5}. Finally, conclusions and limitations are discussed in Section \ref{sec6}.

\section{Related Work}
\label{sec2}
In this section, we will review the methods of single and multi-frame infrared small target detection from the perspective of model and data-driven. Meanwhile, we will introduce related unsupervised learning approaches on two sides.
\subsection{Infrared Small Target Detection (ISTD)}

\textbf{1) Single-frame ISTD: }
As mentioned in Section \ref{sec1}, model-driven single-frame ISTD methods are categorized into background consistency \citep{bai2010tophat}, human-vision-system (HVS) \citep{wei2016mpcm}, and optimization approaches \citep{gao2013ipi,dai2017ript}. Notably, optimization methods dominate as they leverage the low-rank nature of the background and the sparsity of targets within an RPCA model. Gao \etal \citep{gao2013ipi} pioneered this strategy with the infrared patch-image (IPI) scheme in the matrix domain. 
Recognizing that tensors can better represent data structures, Dai \etal \citep{dai2017ript} extended the 2-D matrix into a 3-D tensor construction within the reweighted infrared patch tensor (RIPT) model. This advancement has significantly impacted the ISTD field, prompting researchers to explore various high-dimensional data optimization methods. For instance, Zhang \etal \citep{zhang2019pstnn} introduced a partial sum algorithm to handle nuclear norms and accelerate convergence (PSTNN). Kong \etal \citep{kong2021logtfnn} utilized a fiber-tubal rank to constrain the background. However, these methods often face challenges with complex parameter adjustments.

Based on emerging datasets with ground truth labels, many studies have incorporated DNNs into detecting infrared targets with less parameter fine-tuning. Given the pixel-level importance of small targets, most research treats ISTD as a segmentation task \citep{li2022dnanet,zhang2023agpcnet,sun2023rdian}. However, these methods are often criticized for merely adapting general vision modules without incorporating domain knowledge. In addressing these, several studies \citep{wu2024rpcanet,dai2021alcnet, zhang2022isnet} have attempted to combine HVS or optimization models, achieving relatively interpretable results. Nonetheless, relying solely on spatial information without temporal data can confuse networks when distinguishing between targets and false alarms, leading to occasional miss-detections. Therefore, introducing temporal priors to assist in target allocation is essential.

\textbf{2) Multi-frame ISTD: }
Instead of directly applying optical methods \citep{kwan2020optical} to sequence-based ISTD, several approaches have utilized saliency from the HVS to differentiate targets from the background \citep{du2019stldm}. However, challenges such as kernel selection and unstable performance have limited their broader application. In comparison, optimization methods in the spatial-temporal domain are more preferred,  Liu \etal \citep{liu2021asttv} stacked holistic images into a spatial-temporal tensor and employed an asymmetric total variation (ASTTV) to regularize the background. Luo \etal \citep{luo2024clustering} proposed a clustering and tracking guided system (CTSTC) on 3-D tensors to capture targets. Inspired by patch-based tensor construction. Liu \etal \citep{liu2020small} designed an overlapping tensor structure (STT), and Wang \etal \citep{wang2021npstt} improved it with a non-overlapping strategy (NPSTT). Wu \etal \citep{wu20234d} extended 3-D tensors to 4-D and correlated features using tensor network decomposition tools (4-D TT/TR). Despite these advancements, these methods are constrained by linear tensor/matrix operations, which limit background expression and fail to accurately represent features in complex environments. Additionally, complex tensor structures can become a computational burden as the frame increases.

Several studies \citep{chen2024sstnet,yan2023stdmanet,li2023dtum} have turned to deep learning as multi-frame datasets emerge and deep video detection frameworks advance. However, they face disadvantages similar to single-frame data-driven methods, requiring extensive well-labeled data while only partially incorporating domain knowledge.
Although Deng \etal \citep{deng2024bemst} estimated low-rank background as a prior and accomplished detection with a supervised framework, this complex system still requires labeled ground truths. Therefore, given the limitations of leveraging sequence data in a fully-supervised manner, this study aims to balance domain-specific features with the nonlinear representation capabilities of neural networks using an unsupervised learning approach.

\subsection{Unsupervised Learning for Tensor Representation}
Given the powerful nonlinearity of neural networks, which can effectively solve inverse problems in vision tasks \citep{liu2022deeptensor}, gradual studies leverage deep operations in unsupervised tensor representation processes and can be divided into trained and untrained methods. The first category uses a pre-trained denoiser as a regularizer, acting as a regularizer for the proximal operator in optimization algorithms during each iteration \citep{heide2014flexisp}. 
Based on this, Liu \etal \citep{liu2023wswtnn-pnp} first integrated the deep prior into a multi-frame optimization framework (WSWTNN-PnP) using a trained 3-D denoiser from FFDNet \citep{zhang2018ffdnet} as an auxiliary variable within the alternating direction method of multiplier (ADMM) solver. However, the complexity and sensitivity of the denoiser limit the further improvement of this plug-and-play paradigm.

In contrast, untrained algorithms use DNNs as feature representation tools rather than mapping them to human-made labels and have been adopted in various tensor completion tasks. Luo \etal explored its use in tensor reconstruction and extended it with different decomposition methods \citep{luo2022sstnn} using CNNs.
This inspired Zhang \etal \citep{zhang20233dstpm} to develop a 3DSTPM structure for ISTD, however, it only focuses on representing backgrounds without the targets and consumes massive parameters due to the use of plain 3DCNN. Instead of using CNNs, Luo \etal \citep{luo2023lrtfr} adopted a multi-layer perceptron (MLP) to represent tensors with low-rank aware functions. This INR-based approach has fewer local biases and better preserves complex signals with relatively low parameter size \citep{molaei2023implicit}.
Moreover, recognizing that low rankness alone is insufficient to explore the local correlation of data, Luo \etal \citep{luo2024neurtv} proposed a continuous total variation on low-rank functions, resulting in more efficient regularization. However, this method only considers high-dimensional situations with static objects, such as multispectral/hyperspectral imaging (MSI and HSI), and lacks temporal information representation. Therefore, further research on INR’s application in spatial-temporal contexts, especially for multi-frame ISTD tasks, is highly needed.

\section{Notations and Preliminaries}
\label{sec3}
In our study, scalars, vectors, matrices, and tensors are labeled as $x$, $\textbf{x}$, $\textbf{X}$, and $\mathcal{X}$, respectively. The $i$-th element of $\textbf{x}$ is labeled as $\textbf{x}_{(i)}$. For a differentiable multivariate function $f: \mathbb{R}^{N} \rightarrow \mathbb{R}$, the gradient operator is denoted by $\nabla f(\textbf{x}) = (\frac{\partial f(\textbf{x})}{\partial \textbf{x}_{1}}, \frac{\partial f(\textbf{x})}{\partial \textbf{x}_{2}}, \cdots, \frac{\partial f(\textbf{x})}{\partial \textbf{x}_{N}})^{T}$, which returns a vector containing partial derivatives along the $N$ dimensions. We denote the partial derivative along the $d$-th dimension as $\nabla_{d} f(\textbf{x}) = \frac{\partial f(\textbf{x})}{\partial \textbf{x}_{d}}$. For a tensor $\mathcal{X}$, its unfolding operator along $i$-th mode is defined by $\mathrm{unfold}_i(\cdot): \mathbb{R}^{n_1\times n_2\times n_3} \rightarrow \mathbb{R}^{n_i\times {\prod _{j \ne i}}{n_j}}$, which returns the unfolding matrix $\mathbf{X}^{(i)}:=\mathrm{unfold}_{i}(\mathcal{X})$. The mode-$i (i=1,2,\cdots,t)$ tensor-matrix product is denoted as $\mathcal{X} \times_{i}\textbf{D}= \mathrm{fold}_{i}(\textbf{D}\mathcal{X}^{(i)})$, here $\mathrm{fold}_{i}(\cdot)$ is the inverse operator of $\mathrm{unfold}_{i}(\cdot)$. Following are the brief introductions of Tucker decomposition.

\textbf{Definition 1:} \textit{Tucker Rank \citep{kolda2009tensor}:} For a Tucker factorization in the 3-D tensor domain, for a tensor $\mathcal{C} \in \mathbb{R}^{n_{1}\times n_{2}\times n_{3}}$, its rank is defined by:
\begin{equation}
\setlength{\abovedisplayskip}{-5pt}
\setlength{\belowdisplayskip}{-5pt}
\text{rank}_{\text{Tucker}}(\mathcal{C}) = (r_1,r_2,r_3).
\end{equation}

\textbf{Definition 2:}  \textit{Tucker Decomposition \citep{kolda2009tensor}:} For a three-dimensional tensor $\mathcal{C}$ within the size of $n_1, n_2, n_3$, we can decompose it to a core tensor $\mathcal{G} \in \mathbb{R}^{r_{1}\times r_{2}\times r_{3}}$ and three factor matrices $\textbf{U} \in \mathbb{R}^{n_{1} \times r_{1}}$, $\textbf{V} \in \mathbb{R}^{n_{2} \times r_{2}}$, and $\textbf{W} \in \mathbb{R}^{n_{3} \times r_{3}}$, formulated by:
\begin{equation}
\setlength{\abovedisplayskip}{-5pt}
\setlength{\belowdisplayskip}{-10pt}
\mathcal{C} = \mathcal{G} \times _{1}\textbf{U} \times _{2}\textbf{V} \times _{3}\textbf{W}.
\vspace{-0.4cm}
\end{equation}
\section{Methodology}
\label{sec4}
\subsection{Problem Formulation}
Unlike using patches to construct a tensor, we directly stack consecutive frames $\{\textbf{f}_1,\textbf{f}_2,\cdots,\textbf{f}_{n_3}\} (\textbf{f}_i \in \mathbb{R}^{n_1 \times n_2}, i\!=\!1, 2, \cdots, n_3)$ into a 3-D spatial-temporal tensor $\mathcal{D} \in \mathbb{R}^{n_1 \times n_2 \times n_3}$. We then decompose the original tensor in infrared tasks into background and target tensors $\mathcal{B}, \mathcal{T} \in \mathbb{R}^{n_1 \times n_2 \times n_3}$ as:
\begin{equation}
\setlength{\abovedisplayskip}{-5pt}
\setlength{\belowdisplayskip}{-4pt}
\mathcal{D} = \mathcal{B} + \mathcal{T}.
\vspace{-0.1cm}
\end{equation}
Due to the low-rank nature of infrared backgrounds and the sparsity of targets, detection tasks can be transformed into optimization problems, aiming to minimize the rank of backgrounds and restrict the sparsity of targets:
\begin{equation}
\setlength{\abovedisplayskip}{-3pt}
\setlength{\belowdisplayskip}{-3pt}
\begin{aligned}
 \mathop{\min}\limits_{\mathcal{B},\mathcal{T}} ~~  \text{rank}(\mathcal{B}) + \lambda \left\| {\mathcal{T}} \right\|_{0},  ~~~~s.t. ~~\mathcal{D} = \mathcal{B} + \mathcal{T}.
 \end{aligned}
\end{equation}
where $\lambda$ is a regularization parameter that balances the rank minimization and sparsity constraints. Considering solving $l_0$-norm is the NP-hard, we always adopt tensor RPCA methods \citep{dai2017ript} to transform the two constraints with the nuclear norm $\left\| \cdot \right\|_{*}$ and $l_1$-norm:
\begin{equation}
\setlength{\abovedisplayskip}{-3pt}
\setlength{\belowdisplayskip}{-3pt}
\begin{aligned}
 \mathop{\min}\limits_{\mathcal{B},\mathcal{T}} ~~  &\left\| \mathcal{B} \right\|_{*} + \lambda \left\| {\mathcal{T}} \right\|_{1}, ~~~~s.t. ~~\mathcal{D} = \mathcal{B} + \mathcal{T}.
   \end{aligned}
   \label{pcp}
\end{equation}
However, directly decomposing the background or target element without proper representation is not sufficient, especially under multi-frame scenarios. Thus, this article aims to study a more effective way of representing both elements.
\subsection{Neural-represented Low-Rank Background}
\label{sec:4.2}
To enhance the correlation of the background in ISTD, researchers adopt different unfolding strategies \citep{wu20234d}, but only in the matrix domain. Thus, in balancing the consumption and correlation ability of tensor/matrix,  Tucker factorization may better represent infrared data \citep{liu2023tucker}, where it simultaneously maintains a low-rank property. 

\begin{figure}[t!]
\setlength{\abovecaptionskip}{-6pt}
\centering
\includegraphics[width=0.65\linewidth]{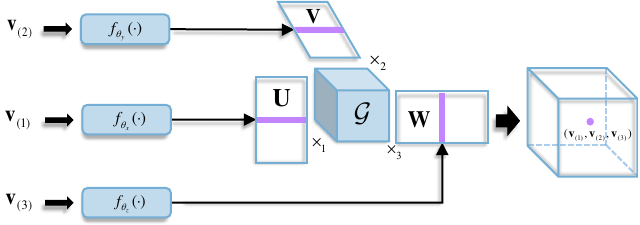}
\caption{Schematic of a tensor function representation method in Eq. (\ref{eq:lrtfr}).}
\label{fig_lrtfr}
\vspace{-0.4cm}
\end{figure}
However, discretely using Tucker decomposition encounters limitations when representing complex tensors with both spatial and temporal information. Therefore, we extend it to the continuous domain, allowing for a more compact representation of multi-dimensional data.

\textbf{1) Tensor Function Representation: }Motivated by \citep{luo2023lrtfr}, we introduce a tensor function to represent the tensor set. We define $f(\cdot): D_{f} = X_f \times Y_f \times Z_f \rightarrow \mathbb{R}$, where the domain in three dimensions is $X_f, Y_f, Z_f \subset \mathbb{R}$. For a tensor function $f(\cdot): D_{f} \rightarrow \mathbb{R}$, the sampled tensor set $S[f]$ is expressed as:
\begin{equation}
\setlength{\abovedisplayskip}{-4pt}
\setlength{\belowdisplayskip}{-5pt}
\begin{aligned}
S[f] = \{ \mathcal{C} \mid \mathcal{C}_{i,j,k} = f(\mathbf{x}_{(i)}, \mathbf{y}_{(j)}, \mathbf{z}_{(k)}), \mathbf{x} \in X^{n_1}_{f}, 
\mathbf{y} \in Y^{n_2}_{f}, \mathbf{z} \in Z^{n_3}_{f}, n_1, n_2, n_3 \in \mathbb{N}_{+} \}.
\end{aligned}
\end{equation}
Here, $\mathbf{x}, \mathbf{y}, \mathbf{z}$ are coordinate vector variables. And the function rank $F$-rank$[f]$ of $f(\cdot)$ is defined as:
\begin{equation}
\setlength{\abovedisplayskip}{-5pt}
\setlength{\belowdisplayskip}{-5pt}
F\text{-rank}[f] = (r_1, r_2, r_3), \text{where} \quad r_i = \sup_{\mathcal{C} \in S[f]} \text{rank}(\mathcal{C}^{(i)}).
\end{equation}

A function $f(\cdot)$ with $F$-rank$[f]$ can be considered a low-rank tensor function, as the Tucker rank of any $\mathcal{C} \in S[f]$ is bounded by $(r_1, r_2, r_3)$. Thus, for any $(v_1, v_2, v_3) \in D_{f}$, we can have a low-rank tensor function factorization:
\begin{equation}
\setlength{\abovedisplayskip}{-3pt}
\setlength{\belowdisplayskip}{-3pt}
f(v_1, v_2, v_3) = \mathcal{G} \times_{1} f_{x}(v_1) \times_{2} f_{y}(v_2) \times_{3} f_{z}(v_3),
\end{equation}
where $f_{x}(\cdot), f_{y}(\cdot), f_{z}(\cdot)$ are three factor functions. For a coordinate variable vector $\bf{v}$ in 3-D, we draw its processing schematic in Fig. \ref{fig_lrtfr} and write its low-rank tensor representation function as:
\begin{equation}
\setlength{\abovedisplayskip}{-2pt}
\setlength{\belowdisplayskip}{-3pt}
[\mathcal{G},f_{x},f_{y},f_{z}](\mathbf{v}) = \mathcal{G}\times _{1}f_{x}(\textbf{v}_{(1)}) \times _{2}f_{y}(\textbf{v}_{(2)}) \times _{3}f_{z}(\textbf{v}_{(3)}).
\label{eq:lrtfr}
\end{equation}

\textbf{2) Neural Implementation: }
Traditionally, these factor functions will be regarded as factor matrices and solved by complex linear calculation \citep{kolda2009tensor}, causing possible information loss and reducing feature correlation. To address this, we employ neural layers with powerful nonlinearity. Specifically, we utilize three MLPs, denoted as $f_{\theta_{x}}(\cdot), f_{\theta_{y}}(\cdot), f_{\theta_{z}}(\cdot)$, with parameters $\theta_{x}, \theta_{y}, \theta_{z}$, to "neuralize" the above three factor functions.
 For instance, given a factor function $f_{{x}}(\cdot)$, we formulated it as:
\begin{equation}
\setlength{\abovedisplayskip}{-2pt}
\setlength{\belowdisplayskip}{-3pt}
f_{{x}}(\cdot) = \mathbf{H}_d(\sigma(\mathbf{H}_{d-1} \cdots \sigma (\mathbf{H}_0 \mathbf{x}))) : X_{f} \rightarrow \mathbb{R}^{r_1}.
\label{eq:inr}
\end{equation}
where $\sigma(\cdot)$ denotes the nonlinear activation function, and $\theta_{x} = \{\mathbf{H}_i\}_{i=0}^{d}$ represents the learnable weight matrices of MLPs, $d$ is neural layer index.

Such a nonlinear representation form can better assist background approximation in complex spatial-temporal scenes that further enhance detection tasks. Thus, we estimate our background tensor with the so-driven method. Here, to suit sequential scenes (height, width, and time), we rename the three functions as ${f_{{\theta{_h}}}},{f_{{\theta{_w}}}},{f_{{\theta{_t}}}}$. The neural-represented low-rank (NLR) background tensor $\mathcal{B}^{\text{NLR}}$ is then defined as:
\begin{equation}
\setlength{\abovedisplayskip}{-3pt}
\setlength{\belowdisplayskip}{-3pt}
\mathcal{B}_{ijk}^{\text{NLR}} = [\mathcal{G};{f_{{\theta{_h}}}},{f_{{\theta{_w}}}},{f_{{\theta{_t}}}}](i,j,k),~~\forall (i,j,k) \in \Gamma,
\label{eq:updateB}
\end{equation}
where $\Gamma\!=\!\{ (i,j,k)|i\!=\!1,2,...,n_1;y = 1,2,...,n_2;k\!=\!1,2,...,n_3\}$. Based on this, Eq. (\ref{pcp}) is rewritten as: 
\begin{equation}
 \vspace{-0.05cm}
\setlength{\abovedisplayskip}{-4pt}
\setlength{\belowdisplayskip}{-5pt}
\begin{aligned}
 \mathop{\min}\limits_{\mathcal{B}^{\text{NLR}},\mathcal{T}}  \left\| {\mathcal{B}^{\text{NLR}}} \right\|_{*} + \lambda \left\| {\mathcal{T}} \right\|_{1}.
 \end{aligned}
 \vspace{-0.05cm}
\label{eq:nlr_pcp}
\end{equation}

In addition, as studied in \citep{sitzmann2020inr}, a periodic sine activation function is more likely to capture a signal's complex structures and details. Thus, instead of using ReLU series activation functions, we adopt sinusoidal representation networks (SIREN) based MLP as our implicit neural network, the activation function is defined as $\sigma(\cdot) = \text{sin}(\cdot)$.

\subsection{Neural 3-D Total Variation Regularization}
\label{sec:4.3}
In most low-level vision tasks, total variation is employed to represent the spatial smoothness of images. However, for higher-level ISTD tasks, we leverage variations in the time field to utilize temporal information, thereby enhancing target detection. Works such as \citep{liu2021asttv} incorporate 3D total variations (3DTV) in discrete operations by:
\begin{equation}
\setlength{\abovedisplayskip}{0pt}
\setlength{\belowdisplayskip}{0pt}
\begin{aligned}
\left\|\mathcal{B}\right\|_{3DTV}&\!=\! \|D_h \mathcal{B}\|_1 + \|D_w \mathcal{B}\|_1 + \|D_t \mathcal{B}\|_1,
\end{aligned}
\label{eq:3dtv_1}
\end{equation}
$D_h$, $D_w$, $D_t$ are discrete difference operators of different orders, where they are defined as: $D_h\mathcal{B}(i,j,k) = \mathcal{B}(i,j,k)\!-\! \mathcal{B}(i,j\!-\!1,k)$, $ D_w\mathcal{B}(i,j,k) = \mathcal{B}(i,j,k)\!-\! \mathcal{B}(i-1,j,k) $, $ D_t\mathcal{B}(i,j,k) = \mathcal{B}(i,j,k)\!-\! \mathcal{B}(i,j,k-1)$), individually.
Based on this, we incorporate the above ${\mathcal{B}^{\text{NLR}}}$ and reformulate Eq. (\ref{eq:nlr_pcp}) as:
\begin{equation}
\setlength{\abovedisplayskip}{-3pt}
\setlength{\belowdisplayskip}{-3pt}
\begin{aligned}
 \mathop{\min}\limits_{\mathcal{B}^{\text{NLR}},\mathcal{T}}  \left\| {\mathcal{B}^{\text{NLR}}} \right\|_{*} + \lambda \left\| {\mathcal{T}} \right\|_{1}\!+\! \phi \left\|  {\mathcal{B}^{\text{NLR}}} \right\|_{3DTV}, 
\end{aligned}
\label{eq:3dtv}
\end{equation}
where $\phi$ is a positive regularization index. However, the classical discrete difference-based TV is criticized for its lack of flexibility and accuracy, as simple subtraction is insufficient for complex data. Therefore, a more flexible operator is needed. Fortunately, by representing the background with DNNs, we can address these issues in a nonlinear manner. Previous works, such as \citep{zhang20233dstpm,luo2022sstnn}, have applied discrete difference operations on DNNs. However, these approaches primarily focus on spatial information or spectral features in still images, neglecting moving features. To fill in this gap, we propose a Neur3DTV regularization, which uses derivatives of DNN outputs to enforce local smoothness and temporal differences in background components. 

Initially, as described in \citep{luo2024neurtv}, given a differentiable function $f_{\Theta}$, a neural total variation (NeurTV) regularization conditioned on $\Theta$ is defined as:
\begin{equation}
\setlength{\abovedisplayskip}{0pt}
\setlength{\belowdisplayskip}{0pt}
\begin{aligned}
{\Psi_{NeurTV}}(\Theta) = \sum\limits_{i=1}^{N}\sum\limits_{\textbf{x}\in\Gamma}\left| \nabla_i {f_\Theta(\textbf{x}) }\right|,
\end{aligned}    
\end{equation}
where $\nabla_i {f_\Theta(\textbf{x})} = \frac{\partial f(\textbf{x})}{\partial \textbf{x}_{(i)}}$ denotes the partial derivative along the $i$-th dimension, with $N$ representing the total number of dimensions. Meanwhile, since we utilize SIREN as the activation function, which maintains its properties in its derivative form (the derivative of $\sin(\cdot)$ is $\cos(\cdot)$), allowing us to capture complex details during regularization.

However, NeurTV only considers spatial constraints, where $N=2$, and does not emphasize temporal information. To address this and in line with conventional 3DTV for small target detection, we formulate a Neur3DTV by:
\begin{equation}
\setlength{\abovedisplayskip}{0pt}
\setlength{\belowdisplayskip}{0pt}
\begin{aligned}
{\Psi_{Neur3DTV}}(\Theta) = \sum\limits_{\textbf{x}\in\Gamma}\left| \nabla_h {f_\Theta(\textbf{x}) }\right| + \sum\limits_{\textbf{x}\in\Gamma}\left| \nabla_w {f_\Theta(\textbf{x}) }\right| + \kappa \cdot \sum\limits_{\textbf{x}\in\Gamma}\left| \nabla_t {f_\Theta(\textbf{x}) }\right|,
\end{aligned}    
\label{eq:neur3dtv}
\end{equation}
where $\kappa$ is the weight coefficient emphasizing temporal information.  
This approach ensures the background preserves local smoothness in continuous-represented DNNs and maintains the significance of temporal regularization.

\subsection{NeurSTT for ISTD}
By plugging the Neur3DTV into the overall ISTD optimization model, our neural-represented spatial-temporal tensor (NeurSTT) model is then built and is written as:
\begin{equation}
\setlength{\abovedisplayskip}{-3pt}
\setlength{\belowdisplayskip}{-3pt}
\begin{aligned}
 \mathop{\min}\limits_{\mathcal{G},{\mathcal{T},{\theta{_h}},{\theta{_w}},{\theta {_t}}}}
\left\| \mathcal{B}^{\text{NLR}} \right\|_{*} \!+\! \lambda{\left\| \mathcal{T} \right\|_{1}}&\!+\! \phi {\Psi_{Neur3DTV}}(\Theta) \\ 
s.t. ~~{\mathcal{B}_{ijk}^{\text{NLR}}} = [\mathcal{G};{f_{{\theta{_h}}}},{f_{{\theta{_w}}}},{f_{{\theta{_t}}}}]&(i,j,k),~~\forall (i,j,k) \in \Gamma.
\end{aligned}    
\end{equation}

For simplicity, we employ an alternating minimization algorithm to address the proposed model, separately handling the background and the target.
For background components:
\begin{equation}
\setlength{\abovedisplayskip}{-3pt}
\setlength{\belowdisplayskip}{-3pt}
\begin{aligned}
 \mathop{\min}\limits_{\mathcal{G},{\theta{_h}},{\theta{_w}},{\theta {_t}}}
 \frac{1}{2} \left\| \mathcal{D} \!-\! \mathcal{B}^{\text{NLR}} \!-\! \mathcal{T} \right\|_{F}^{2} + \left\| \mathcal{B}^{\text{NLR}} \right\|_{*} \!+\! \phi {\Psi_{Neur3DTV}}(\Theta). 
\end{aligned}    
\end{equation}

\begin{algorithm}[t!]
	\caption{Solution Process of NeurSTT for ISTD}
	\label{alg:neurstt}
	\KwIn{Original infrared tensor sequence ${{\cal D}\in \mathbb{R}^{n_1 \times n_2 \times n_3}}$, parameter $\lambda$, $\phi$, down sample rate $r_d$, and $\kappa$.}
	\KwOut{Background tensor ${\cal B}$ and target tensor ${\cal T}$. }  
	\BlankLine
	
	Initialize F-rank $[r_1, r_2, r_3] = [n_1/r_d, n_2/r_d, n_3/r_d]$, core tensor ${\cal G}\in \mathbb{R}^{r_1 \times r_2 \times r_3}$, factor MLPs $f_{\theta_h}, f_{\theta_w}, f_{\theta_t}$, and maximum steps of iteration $k_{max} = 2000$.
	
	\While{\text{not converge} or $k < k_{max}$}{
        Compute the approximate background $\mathcal{B}^{\text{NLR}}$ via Eq. (\ref{eq:updateB});

        Compute the $\Psi_{Neur3DTV}$ via Eq. (\ref{eq:neur3dtv});

        Compute the separated target ${\cal T}$ via Eq. (\ref{eq:updateT});

        Compute the loss in Eq. (\ref{eq:loss_all});

        Update ${\cal G}$ and MLP parameters with the loss using the Adam;
	        
	}
\vspace{-0.1cm}
\end{algorithm}

In traditional solvers, such sub-problems are typically addressed using singular value thresholding-based strategies \citep{kolda2009tensor}. However, these approaches are complex and may lose essential tensor features. Instead, as discussed in Section \ref{sec:4.2}, we approximate the background tensor by neural layers with learnable parameters, their powerful nonlinearity can assist in complex feature capturing. Similar to \citep{luo2023lrtfr}, we update these parameters using the Adam gradient descent method \citep{kingma2014adam} within set iterations. Meanwhile, a critical concern is how to regularize this in the learning schema. Motivated by \citep{zhang20233dstpm,luo2022sstnn}, and inheriting the merit of traditional methods, we utilize nuclear norm in time series as a sub-loss function to constrain the background, the $\mathcal{L}_{\text{Nuc}}$ is defined as:
\begin{equation}
\setlength{\abovedisplayskip}{-1pt}
\setlength{\belowdisplayskip}{-1pt}
\begin{aligned}
\mathcal{L}_{\text{Nuc}} = \sum\limits_{k = 1}^{n_3} {\left\| {\mathcal{B}_{k}^{\text{NLR}}} \right\|_{*}}.
\end{aligned}    
\end{equation}

For $\Psi_{Neur3DTV}$, we explicitly calculate derivatives of the SIREN functions (as derived in Section \ref{4.5}) and use a general $l_1$-norm for loss regularization:
\begin{equation}
\setlength{\abovedisplayskip}{-4pt}
\setlength{\belowdisplayskip}{-3pt}
\begin{aligned}
\mathcal{L}_{\text{Neur3DTV}} = \phi \left\|{\Psi_{Neur3DTV}}(\Theta)\right\|_{1}.
\end{aligned}    
\end{equation}

For the target, its sub-problem is formulated as: 
\begin{equation}
\setlength{\abovedisplayskip}{-1pt}
\setlength{\belowdisplayskip}{-1pt}
\begin{aligned}
\mathop{\min}\limits_{{\mathcal{T}}}
\frac{1}{2} \left\| \mathcal{D} - \mathcal{B}^{\text{NLR}} - \mathcal{T} \right\|_{F}^{2} + \lambda\left\|\mathcal{T}\right\|_{1}.
\end{aligned}    
\end{equation}
Unlike \citep{zhang20233dstpm}, which treats $\mathcal{T}$ as the result of subtracting the background from the original tensor, we solve the $\mathcal{T}$ using a soft-thresholding method, formulated by:
\begin{equation}
\setlength{\abovedisplayskip}{-5pt}
\setlength{\belowdisplayskip}{-3pt}
    \mathcal{T} = \mathcal{ST}_{\frac{\mathcal{\lambda}}{2}}(\mathcal{D} \!-\! \mathcal{B}^{\text{NLR}}),
\label{eq:updateT}
\end{equation}
$\mathcal{ST}(\cdot)$ indicates the soft-thresholding operator \citep{ma2011fixed} ($\mathcal{ST}_{v}(\cdot)=\text{sgn}(\cdot)\max\{|\cdot|\!-\!v, 0\}$). To constrain the target in the learning process, we also introduce an $l_1$-norm as a loss element ($\mathcal{L}_{\mathcal{T}}$) to regularize the target representation:
\begin{equation}
\setlength{\abovedisplayskip}{-5pt}
\setlength{\belowdisplayskip}{-3pt}
\begin{aligned}
\mathcal{L}_{\mathcal{T}} = \left\|\mathcal{T}\right\|_{1}.
\end{aligned}    
\label{eq_lossT}
\end{equation}

We collect the overall loss function $\mathcal{L}_{\text{All}}$ as follows:
\begin{equation}
\setlength{\abovedisplayskip}{-4pt}
\setlength{\belowdisplayskip}{-3pt}
\begin{aligned}
\mathcal{L}_{\text{All}} = \mathcal{L}_{\text{Nuc}} + \mathcal{L}_{\mathcal{T}} + \mathcal{L}_{\text{Neur3DTV}}.
\end{aligned} 
\label{eq:loss_all}
\end{equation}

\begin{figure*}[t!]
\vspace{-0.1cm}
\setlength{\abovecaptionskip}{-3pt}
\setlength{\belowcaptionskip}{0pt}
\centering
\includegraphics[width=0.98\linewidth]{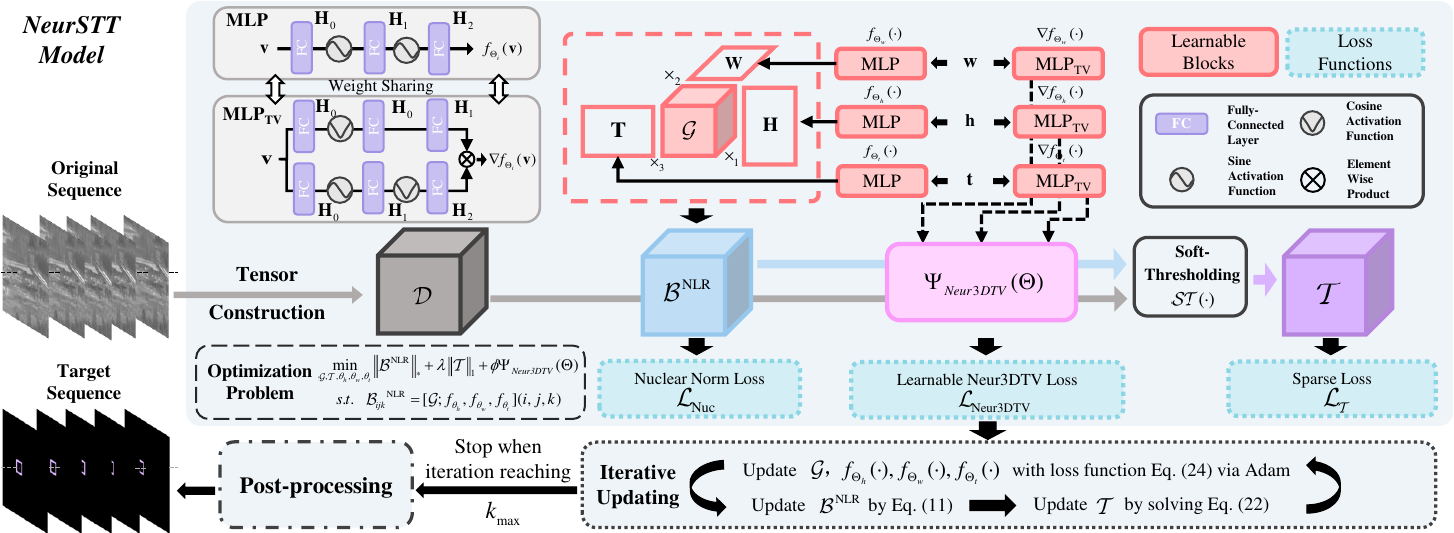}
\caption{
Overall procedure of NeurSTT model for ISTD. The background spatial-temporal tensor is represented by a neural function with a low-rank prior, $\mathcal{B}^{\text{NLR}}$, followed by 3-D neural total variation regularization, $\Psi_{Neur3DTV}$. The target tensor is determined using a soft-thresholding operator. Each module has a specific loss constraint, combined into an overall loss function, with parameters updated using deep strategies.}
\label{fig_overall}
\vspace{-0.3cm}
\end{figure*}

The overall procedure of the NeurSTT model is outlined in Algorithm \ref{alg:neurstt} and drawn in Fig. \ref{fig_overall}. Here, similar to \citep{zhang20233dstpm}, we post-process the result tensor with a general tensor-to-sequence operation and use an adaptive threshold procedure to generate the final target sequences.

\subsection{Network Details}
\label{4.5}
NeurSTT has a set of learnable parameters in updating processes, denoted as $\Theta = \{{\mathcal{G};{f_{{\theta{_h}}}},{f_{{\theta{_w}}}},{f_{{\theta{_t}}}}}\}$. The core tensor $\mathcal{G}$ is a learnable parameter tensor, initialized with a uniform distribution. For the factor matrices, we initialize them with corresponding coordinates with $\textbf{h} \in \mathbb{R}^{n_1}$, $\textbf{w} \in \mathbb{R}^{n_2}$, $\textbf{t} \in \mathbb{R}^{n_3}$. Given the example of $\textbf{h}$, the channel is increased by neural layers, and its output is represented by:
\begin{equation}
\setlength{\abovedisplayskip}{-2pt}
\setlength{\belowdisplayskip}{-2pt}
\begin{aligned}
\textbf{H} = f_{\theta_h}({\textbf{h}}) = \textbf{H}_d(\text{sin}(\textbf{H}_{d-1}\cdots\text{sin}(\textbf{H}_0\cdot\textbf{h})))),
\end{aligned}    
\end{equation}
where the output $\textbf{H}$ is in $ \mathbb{R}^{r_1 \times n_1}$, and the weight matrices $\textbf{H}_0 \in \mathbb{R}^{1 \times c_i}$, $\textbf{H}_{i \ne 0,n} \in \mathbb{R}^{c_i \times c_i}$, $\textbf{H}_n \in \mathbb{R}^{c_i \times r_1}$ will be initialized as described in \citep{sitzmann2020inr} ($\textbf{W}  \in \mathbb{R}^{r_2 \times n_2}$ and $\textbf{T}\in \mathbb{R}^{r_3 \times n_3}$ follow the similar procedure). The channel indexes are set as $c_i = n_i ~(i=1,2,3)$ and the uniform value is set as $std = 5$. For the MLP networks in the background approximation part, we show the network example in Fig. \ref{fig_overall} when $d = 2$.

For another learnable block Neur3DTV, we denote its sub-network as MLP$_{\text{TV}}$. Given the example when $d = 2$, we write its derivative as follows (details can be found in \ref{appendix}):
\begin{equation}
\setlength{\abovedisplayskip}{-2pt}
\setlength{\belowdisplayskip}{-2pt}
\begin{aligned}
\nabla_h {f_\Theta(\textbf{h})} \!=\!\textbf{H}_2 \cos(\textbf{H}_1 \sin(\textbf{H}_0 \cdot \textbf{h}))\!\cdot\! \textbf{H}_1\cos(\textbf{H}_0\!\cdot\!\textbf{h})\!\cdot\!\textbf{H}_0.
\end{aligned}    
\end{equation}

Based on this, we shape the network as demonstrated in Fig. \ref{fig_overall}. Here, $\{\mathbf{H}_{i}\}^{d}_{i=0}$ of MLP$_{\text{TV}}$ represent the shared weights in MLP networks, which indicates that no additional parameters are required, demonstrating the efficacy of the proposed method.

\section{Experiment and Result}
\label{sec5}
In this section, we introduce the datasets, baseline methods, and evaluation metrics. We then conduct ablation studies within our NeurSTT, compare our results with state-of-the-art (SOTA) methods, and discuss the model’s robustness in noisy environments.

\subsection{Detail Implementation}
We use two public datasets \citep{sun2023rdian,hui2019atr} for evaluation. From the \citep{hui2019atr} dataset, we select six complex real-world scenes, and from the \citep{sun2023rdian} dataset, we choose three extremely noisy and complex scenes. Detailed configurations and descriptions of these sequences are provided in Table \ref{tab:1}.

The learning rate and weight decay index are individually set to 0.0005 and 0.01 during the experiment. Our method is implemented in PyTorch and runs on an Intel Xeon Platinum 8222CL CPU and Nvidia GeForce 3090 GPU. Experiments on traditional methods are conducted using Matlab R2021b, the code will be available at \url{https://github.com/fengyiwu98/NeurSTT}.

To fairly evaluate our method, we select several SOTA methods as our comparison baselines. These include single-frame-based methods: MPCM \citep{wei2016mpcm}, IPI \citep{gao2013ipi}, PSTNN \citep{zhang2019pstnn}, RPCANet \citep{wu2024rpcanet}, and RDIAN \citep{sun2023rdian}; and multi-frame-based methods: STT \citep{liu2020small}, ASTTV-NTLA \citep{liu2021asttv}, NPSTT \citep{wang2021npstt}, NFTDGSTV \citep{liu2023tucker}, CTSTC \citep{luo2024clustering}, WSWTNN-PnP \citep{liu2023wswtnn-pnp}, and 3DSTPM \citep{zhang20233dstpm}. Their detailed configurations are listed in Table \ref{table_para}.
\begin{table*}[]
\setlength\abovecaptionskip{0pt}
\setlength\belowcaptionskip{0pt}
\caption{Representative samples and corresponding descriptions of nine sequences from different datasets.}
\centering
{\tiny
\resizebox{0.9\linewidth}{!}{
\begin{tabularx}{0.9\linewidth}{>{\centering\arraybackslash}p{0.3cm}
        >{\centering\arraybackslash}p{1.4cm}
        >{\centering\arraybackslash}p{0.5cm}
         >{\centering\arraybackslash}p{1.2cm}
         >{\raggedright\arraybackslash}p{8cm}}
\hline
Seq. &Example& Num. & Image size & Detail Description \\\hline
\multirow{2}{*}{1}&\begin{minipage}[b]{1.5cm}\centering \adjustbox{valign=c, width=1.5cm}{\includegraphics{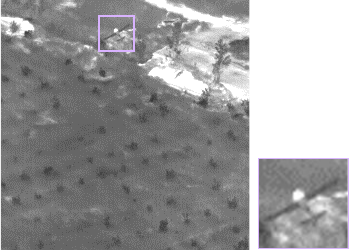}} \end{minipage}& \multirow{2}{*}{80} & \multirow{2}{*}{\textbf{$256\times256$ }} &  $\bullet$ Fast-moving tiny targets with a relatively slow-moving background;\newline $\bullet$ Noisy clusters include buildings, bushes, and have a low SCR.\\
\multirow{2}{*}{2}&\begin{minipage}[b]{1.5cm}\centering \adjustbox{valign=c, width=1.5cm}{\includegraphics{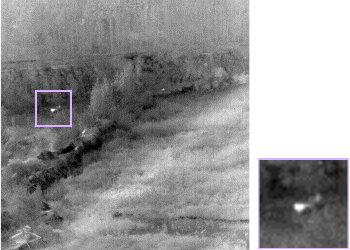}} \end{minipage}& \multirow{2}{*}{80} & \multirow{2}{*}{\textbf{$256\times256$ }} &  $\bullet$ Relatively large, fast-moving targets with background changes;\newline $\bullet$ Ground-forest-sky scene with strong clusters.   \\
\multirow{2}{*}{3}&\begin{minipage}[b]{1.5cm}\centering \adjustbox{valign=c, width=1.5cm}{\includegraphics{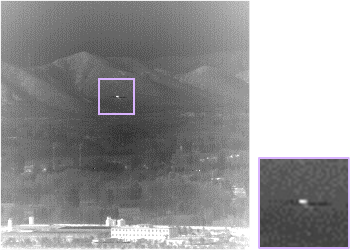}} \end{minipage}& \multirow{2}{*}{80} & \multirow{2}{*}{\textbf{$256\times256$ }} & $\bullet$ Tiny unmanned aerial vehicle with a relatively shaped head; \newline $\bullet$  Complex scenario with bright buildings, mountains, and point noise.        \\
\multirow{2}{*}{4}&\begin{minipage}[b]{1.5cm}\centering \adjustbox{valign=c, width=1.5cm}{\includegraphics{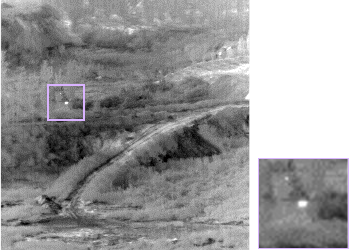}} \end{minipage}& \multirow{2}{*}{80} & \multirow{2}{*}{\textbf{$256\times256$ }} &  $\bullet$ Fast-moving dim and small target within ground scenes;\newline $\bullet$ False alarms include bright edges, highlight tiny points and bushes. \\
\multirow{2}{*}{5}&\begin{minipage}[b]{1.5cm}\centering \adjustbox{valign=c, width=1.5cm}{\includegraphics{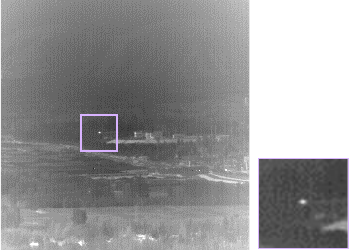}} \end{minipage}& \multirow{2}{*}{80} & \multirow{2}{*}{\textbf{$256\times256$ }} &  $\bullet$ Relative shaped moving point from sky to complex ground; \newline $\bullet$ Massive distributed buildings with bush-filled foreground.   \\
\multirow{2}{*}{6}&\begin{minipage}[b]{1.5cm}\centering \adjustbox{valign=c, width=1.5cm}{\includegraphics{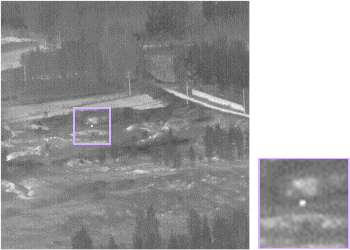}} \end{minipage}& \multirow{2}{*}{80} & \multirow{2}{*}{\textbf{$256\times256$ }} &  $\bullet$ Small-sized target with large range movement;\newline $\bullet$ Massive target-like sparse cluster, lamps, and road's marginal cluster    \\
\multirow{2}{*}{7}&\begin{minipage}[b]{1.5cm}\centering \adjustbox{valign=c, width=1.5cm}{\includegraphics{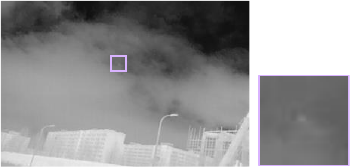}} \end{minipage}& \multirow{2}{*}{80} & \multirow{2}{*}{\textbf{$720\times480$ }} & $\bullet$ Fast-moving tiny and dim target in the building-sky scenes;\newline $\bullet$  False alarms include target-like street lamps and noisy clouds.\\
\multirow{2}{*}{8}&\begin{minipage}[b]{1.5cm}\centering \adjustbox{valign=c, width=1.5cm}{\includegraphics{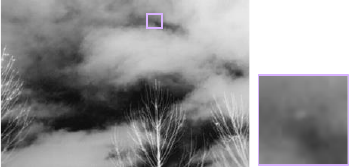}} \end{minipage}& \multirow{2}{*}{80} & \multirow{2}{*}{\textbf{$720\times480$ }} & $\bullet$ Ground-to-sky environment with widely distributed tree branches;\newline $\bullet$ Targets are small and dim, moving from clouds to clear sky.  \\
\multirow{2}{*}{9}&\begin{minipage}[b]{1.5cm}\centering \adjustbox{valign=c, width=1.5cm}{\includegraphics{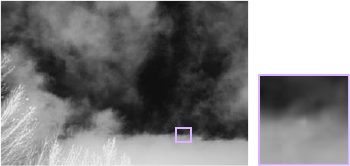}} \end{minipage}& \multirow{2}{*}{80} & \multirow{2}{*}{\textbf{$720\times480$ }} &  $\bullet$ 
Extremely small-sized target with wide range movement;\newline $\bullet$ Massive target-like sparse leaves and edge clusters of clouds. \\
\hline
\end{tabularx}}}
\label{tab:1}
\end{table*}

\subsection{Metrics}
For a comprehensive evaluation, we adopt two categories of evaluation metrics in this study: target-level and pixel-level.

\begin{table*}[t!]
\vspace{-0.2cm}
\setlength{\abovecaptionskip}{-1pt}
\setlength{\belowcaptionskip}{1pt}
\caption{Parameter settings of compared methods.}
\centering
\resizebox{\linewidth}{!}{
\begin{tabular}{cc}
\hline
Methods&Parameters  \\
\midrule
MPCM \citep{wei2016mpcm}&Mean filter index: $i_n\!\times\!i_n,\ i_n: 3,5,7,9$\\
IPI \citep{gao2013ipi}&Patch index: $i_1\!\times\!i_2: 30 \times 30$, sliding index: 10, $\lambda: {\raise0.7ex\hbox{$1$} \!\mathord{\left/
 {\vphantom {1 {\sqrt {\min ({i_1},{i_2})} }}}\right.\kern-\nulldelimiterspace}
\!\lower0.7ex\hbox{${\sqrt {\min ({i_1},{i_2})} }$}}$, \ $\varepsilon: 10^{-7} $ \\
PSTNN \citep{zhang2019pstnn}&Patch index: $i_1\!\times\!i_2: 40\!\times\!40$, sliding index: 40, $\lambda: {\raise0.7ex\hbox{$1$} \!\mathord{\left/
 {\vphantom {1 {\sqrt {\min ({i_1},{i_2})} }}}\right.\kern-\nulldelimiterspace}
\!\lower0.7ex\hbox{${\sqrt {\max ({i_1},{i_2})*n_3} }$}}$, \ $\varepsilon: 10^{-7} $ \\
STT \citep{liu2020small}&Patch index: $i_1\!\times\!i_2: 50\!\times\!50$, sliding index: 40, $\lambda: {\raise0.7ex\hbox{$1$} \!\mathord{\left/
 {\vphantom {1 {\sqrt {\min ({i_1},{i_2})} }}}\right.\kern-\nulldelimiterspace}
\!\lower0.7ex\hbox{${\sqrt {\max ({i_1},{i_2})*i_3} }$}}$, \ $\varepsilon: 10^{-7} $ \\
ASTTV-NTLA \citep{liu2021asttv}&Temporal index: $L: 3$, $H: 6$, $\lambda_{tv}: 0.005$, $\lambda_{3}: 100$, $\lambda: {\raise0.7ex\hbox{$H$} \!\mathord{\left/
 {\vphantom {1 {\sqrt {\min ({M},{N})} }}}\right.\kern-\nulldelimiterspace}
\!\lower0.7ex\hbox{${\sqrt {\max ({i_1},{i_2})*i_3} }$}}$, \ $\varepsilon: 10^{-6} $ \\
NPSTT \citep{wang2021npstt}&Patch index: $i_1\!\times\!i_2: 70\!\times\!70$, Temporal index: $p: 3$, $\lambda: {\raise0.7ex\hbox{$1$} \!\mathord{\left/
 {\vphantom {0.5 {\sqrt {\min ({m_1},{n_1})} }}}\right.\kern-\nulldelimiterspace}
\!\lower0.7ex\hbox{${\sqrt {\max ({i_1},{i_2})*i_3} }$}}$, \ $\alpha: 10^{-3} $ \\
NFTDGSTV \citep{liu2023tucker}&Temporal index: $L: 3$, $C: 5$, $H:4$, $\lambda_1:  0.01$, $\lambda_s:  0.001$, $\lambda: {\raise0.7ex\hbox{$H$} \!\mathord{\left/
 {\vphantom {0.5 {\sqrt {\min ({m_1},{n_1})} }}}\right.\kern-\nulldelimiterspace}
\!\lower0.7ex\hbox{${\sqrt {\max ({i_1},{i_2})*i_3} }$}}$, \ $\epsilon: 10^{-6} $ \\
CTSTC \citep{luo2024clustering}&Temporal index: $f: 3$, $\lambda_L: 30$, $\mu:  0.01$, $\lambda_s:  0.001$, $\lambda_1: {\raise0.7ex\hbox{$\lambda_L$} \!\mathord{\left/
 {\vphantom {0.5 {\sqrt {\min ({m_1},{n_1})} }}}\right.\kern-\nulldelimiterspace}
\!\lower0.7ex\hbox{${\sqrt {\max ({i_1},{i_2})*i_3} }$}}$, $\lambda_2$: $80\lambda_1$, $\lambda_3$: $10\lambda_1$, $\tau : 0.5$ $\sigma: 10^{-4} $ \\
WSWTNN-PnP \citep{liu2023wswtnn-pnp}&Patch index:  $i_1\!\times\!i_2: 30\!\times\!30$, $\lambda: {\raise0.7ex\hbox{$2.4$} \!\mathord{\left/
 {\vphantom {1 {\sqrt {\min ({M},{N})} }}}\right.\kern-\nulldelimiterspace}
\!\lower0.7ex\hbox{${\sqrt {\max ({i_1},{i_2})*i_3} }$}}$, $\tau: 10^{-6} $\\
3DSTPM \citep{zhang20233dstpm}&Temporal index: $l: 80$, ${\lambda}: 0.2$, ${\lambda _2}: 0.65$, $n_4: 160$, $m: 2$, $t_{max}: 3000$\\
Ours&Temporal index:  $L: 80$, ${\lambda}: 0.2$, $\phi: 5\times10^{-5}$, $\kappa: 100$, $k_{max}: 2000$\\
\hline
\end{tabular}}
\vspace{-0.2cm}
\label{table_para}
\end{table*}

\textbf{1) Target-level:}
For the target-level evaluation, we use the three-dimensional receiver operating characteristics (3-D ROC) \citep{chang20203droc}. The 3-D ROC coordinates are the true positive rate (TPR), false positive rate (FPR), and $\tau$ for the threshold. These are paired in different planes as $\text{ROC}_{(\text{TPR},\text{FPR})}$, $\text{ROC}_{(\tau,\text{TPR})}$, and $\text{ROC}_{(\tau,\text{FPR})}$. The areas under the curves (AUC) are denoted as $\text{AUC}_{(\text{TPR},\text{FPR})}$, $\text{AUC}_{(\tau,\text{TPR})}$, and $\text{AUC}_{(\tau,\text{FPR})}$, respectively. Additionally, we evaluate the algorithm's performance on the following AUC series: background suppression (BS), target detection (TD), and their combination, the overall detection probability (ODP):
\begin{equation}
\setlength{\abovedisplayskip}{-4pt}
\setlength{\belowdisplayskip}{-4pt}
    \text{AUC}_{\text{BS}} = \text{AUC}_{(\text{TPR},\text{FPR})} - \text{AUC}_{(\tau,\text{FPR})},
\end{equation}
\begin{equation}
\setlength{\abovedisplayskip}{-4pt}
\setlength{\belowdisplayskip}{-4pt}
    \text{AUC}_{\text{TD}} = \text{AUC}_{(\text{TPR},\text{FPR})} + \text{AUC}_{(\tau,\text{TPR})},
\end{equation}
\begin{equation}
\setlength{\abovedisplayskip}{-4pt}
\setlength{\belowdisplayskip}{-4pt}
    \text{AUC}_{\text{TDBS}} = \text{AUC}_{(\tau,\text{TPR})} - \text{AUC}_{(\tau,\text{FPR})},
\end{equation}
\begin{equation}
\setlength{\abovedisplayskip}{-4pt}
\setlength{\belowdisplayskip}{-4pt}
    \text{AUC}_{\text{ODP}} =  \text{AUC}_{(\text{TPR},\text{FPR})} + \text{AUC}_{(\tau,\text{TPR})} - \text{AUC}_{(\tau,\text{FPR})}.
\end{equation}
Additionally, considering that false alarms are mostly caused by residuals, we introduce the signal-to-noise probability ratio (SNPR) as a metric for evaluation.
\begin{equation}
\setlength{\abovedisplayskip}{-4pt}
\setlength{\belowdisplayskip}{-4pt}
    \text{AUC}_{\text{SNPR}} =  \text{AUC}_{(\tau,\text{TPR})} /\text{AUC}_{(\tau,\text{FPR})}.
\end{equation}

\textbf{2) Pixel-level:}
Thanks to the labeling efforts by \citep{sun2023rdian,hui2019atr}, we can further evaluate our performance using pixel-level metrics: intersection over union ($IoU$) and F-measure ($F_1$). The $IoU$ is defined as:
\begin{equation}
\setlength{\abovedisplayskip}{-2pt}
\setlength{\belowdisplayskip}{-2pt}
    IoU = \text{\#Overlapping Area} / \text{\#Union Area}.
\end{equation}

The $F_1$ score is defined as:
\begin{equation}
\setlength{\abovedisplayskip}{-2pt}
\setlength{\belowdisplayskip}{-2pt}
    F_1 = 2 \times \text{Precision} \times \text{Recall} / \text{Precision} + \text{Recall}.
\end{equation}
where $\text{Precision} = \frac{\text{TP}}{\text{TP+FP}}$ and $\text{Recall} = \frac{\text{TP}}{\text{TP+FN}}$ as defined in \citep{li2022dnanet}. For pixel evaluations, we use a threshold value of 0.5 for traditional methods and 0 for RDIAN and RPCANet.

\subsection{Ablation Studies}
In this section, we first discuss the effect of essential parameters in optimization models. Next, we examine the impact of different network structures and configurations on our NeurSTT model. Notably, we use the average values of two pixel-level metrics, $IoU$ and $F_1$, across six sequences from \citep{hui2019atr} for this analysis.
\begin{table}[t!]
\setlength\abovecaptionskip{0pt}
\setlength\belowcaptionskip{0pt}
    \caption{Ablation Studies on different hyperparameters in NeurSTT on all six sequences from \citep{hui2019atr} in averaged $IoU$ and $F_1$ (\%).}
    \centering 
    {\tiny
    \begin{tabular}{cccccccc}
    \hline
      Config.   & $L$ & $\lambda$ & $\phi$ & $\kappa$  & $k_{max}$& $IoU$ &$F_1 $\\\hline
        \multirow{4}{*}{I} &  10& 0.20 &  $5e^{-5}$ &100  & 2000 &51.90  &66.36\\
        &  20& 0.20 &  $5e^{-5}$ &100  & 2000 &66.35&79.23 \\
        &  40& 0.20 &  $5e^{-5}$ &100  & 2000 &78.21&87.64 \\
        &  80& 0.20 &  $5e^{-5}$ &100  & 2000 &\textbf{81.44}&\textbf{89.68} \\\hline
        \multirow{4}{*}{II} &  80& 0.05 &$5e^{-5}$  & 100 &2000  &38.70&51.94 \\
        &  80& 0.10 & $5e^{-5}$ &100  & 2000 &52.77&66.41 \\
        &  80& 0.15 & $5e^{-5}$ &100  & 2000 &64.36&76.16\\
        &  80& 0.20 & $5e^{-5}$ &100  & 2000 &\textbf{81.44}&\textbf{89.68}\\
        &  80& 0.25 & $5e^{-5}$ &100  & 2000 &75.66&85.33\\\hline
        \multirow{4}{*}{III} &  80& 0.20 &  $1e^{-5}$ & 100 & 2000 &80.12&88.86 \\
        &  80& 0.20 & $5e^{-5}$ & 100 & 2000 &\textbf{81.44}&\textbf{89.68} \\
        &  80& 0.20 & $1e^{-4}$ &100  & 2000 &77.00&86.83 \\
        &  80& 0.20 & $5e^{-4}$ &100  & 2000 &78.13&87.54\\\hline
        \multirow{4}{*}{IV} &  80& 0.20 &  $5e^{-5}$& 1 &2000  &77.62&87.08 \\
        &  80& 0.20 & $5e^{-5}$ &  10& 2000 &71.25&82.43 \\
        &  80& 0.20 & $5e^{-5}$ &  100& 2000 &\textbf{81.44}&\textbf{89.68}\\
        &  80& 0.20 & $5e^{-5}$ &  1000& 2000 &73.27&83.47 \\\hline
        \multirow{4}{*}{V} &  80& 0.20 & $5e^{-5}$ & 100 &  1000&68.22&78.13 \\
        &  80& 0.20 & $5e^{-5}$ &100  & 1500 &75.83&86.03\\
        &  80& 0.20 & $5e^{-5}$ &100  & 2000&\textbf{81.44}&\textbf{89.68} \\
        &  80& 0.20 & $5e^{-5}$ & 100 &  2500&77.96&86.50 \\
        &  80& 0.20 & $5e^{-5}$ & 100 &  3000&77.57&86.97 \\
        \hline
    \end{tabular}}
    \label{tab:parameter}
\vspace{-0.3cm}
\end{table}

\textbf{1) Impact of Index $L$:} The temporal index $L$ is crucial for capturing temporal information, as setting it too low or too high can affect sufficiency. Additionally, $L$ influences the running time of our algorithm. Therefore, we study the impact of different $L$ values (10, 20, 40, and 80) in Config. I from Table \ref{tab:parameter}. Our findings indicate that with the same parameter settings, detection performance improves as $L$ increases. Given the doubled running time and relatively poor performance when $L=40$, we set $L=80$.

\textbf{2) Impact of Index $\lambda$:} $\lambda$ is the trade-off coefficient between background and targets. Finding an appropriate $\lambda$ to minimize false alarms while preserving targets is crucial. We evaluate different $\lambda$ values (0.05, 0.10, 0.15, 0.20, and 0.25) in Config. II from Table \ref{tab:parameter}. Our findings show that when $\lambda = 0.05$, there is a significant increase in missed detections. In contrast, when $\lambda = 0.20$ or $0.25$, the performance is satisfactory. Thus, we select $\lambda = 0.20$ for our setting.

\textbf{3) Impact of Index $\phi$ and $\kappa$:} The trade-off factor $\phi$ determines the influence of our Neur3DTV. We investigate how this index affects the background's spatial-temporal feature representation and, consequently, the detection task. We evaluate $\phi$ values of $1e^{\!-\!5}$, $5e^{\!-\!5}$, $1e^{\!-\!4}$, and $5e^{\!-\!4}$ in Config. III from Table \ref{tab:parameter}. Our results show that, due to the robustness of our network, all settings yield good results, with $\phi\!=\! 1e^{-5}$ performing the best. Thus, we select $\phi\!=\! 1e^{\!-\!5}$ as trade-off value.

Additionally, we emphasize the temporal component in our Neur3DTV by adding the temporal derivative and assigning it a weight parameter $\kappa$. We evaluate $\kappa$ values of 1, 10, 100, and 1000 in Config. IV from Table \ref{tab:parameter}. All parameter settings perform well, exceeding 70 in $IoU$. However, when $\kappa = 100$, our model achieves the best performance, so we select this value as our temporal trade-off parameter.

\textbf{4) Stop Iteration $k_{max}$:} In our learning framework, the stopping iteration sets the convergence criterion. The number of iterations ($k_{max}$) can influence both the convergence of the proposed model and the detection effectiveness of our network, as well as the training time. We study $k_{max}$ values of 1000, 1500, 2000, 2500, and 3000 in Config. V from Table \ref{tab:parameter}. Our results show that the efficacy of our model peaks at $k_{max} = 2000$. Although $k_{max} = 2500$ and $3000$ have similar performance, considering the time consumption, we choose $k_{max} = 2000$ as our maximum iteration number.
\begin{table}[t!]
\setlength\abovecaptionskip{0pt}
\setlength\belowcaptionskip{0pt}
    \caption{Studies on the downsampling setting, as well as corresponding parameter amount in NeurSTT on all six sequences from \citep{hui2019atr} in averaged $IoU$ and $F_1$ (\%).}
    \centering 
    {\scriptsize
    \begin{tabular}{p{2.2cm}<{\raggedright}p{1.2cm}<{\centering}p{1cm}<{\centering}p{1cm}<{\centering}p{1cm}<{\centering}}
    \hline
        Methods&Params&Runtime& $IoU$ $\uparrow$ & $F_1$ $\uparrow$\\\hline
        NeurSTT (Full)& 5.330 M&0.939 s&45.02 &6094\\
        NeurSTT ($1/2$)& 0.885 M& 0.924 s&76.85  & 86.83\\
        \rowcolor{gray!20}NeurSTT ($1/4$)& 0.317 M&0.866 s&\textbf{81.44}&\textbf{89.68}\\
        NeurSTT ($1/8$)& 0.216 M&0.804 s&60.24&73.71\\
        3DSTPM& 5.269 M&1.696 s&68.33&80.70  \\
        \hline
    \end{tabular}}
        \vspace{-0.3cm}
    \label{tab:downsample}
\end{table}

\textbf{5) Impact of Downsampling Rate $r_d$ in Tucker Rank Settings:} As we use Tucker decomposition as the fundamental backbone of our network, determining the appropriate down-sampling rate $r_d$ is crucial. Full usage might cause feature redundancy, while too little might lead to insufficient background recovery. We evaluate different $r_d$ of 1 (Full), $1/2$, $1/4$, and $1/8$ in rank settings. In addition to our previous discussions, we introduce two computational metrics: the number of parameters used and the running time for different rank settings. We also compare our results with the similarly categorized 3DSTPM as a control subject.
As shown in Table \ref{tab:downsample}, under the same parameter setting, as $r_d$ decreases, the performance of NeurSTT improves, reaching its peak when $r_d = 1/4$. Although the running time is minimized at $1/8$, its corresponding performance is relatively poor. Meanwhile, when $r_d = 1/4$, NeurSTT is 48.9\% faster, $16.6\times$ lighter, and $19.19\%$ higher in $IoU$ than baseline 3DSTPM. Therefore, we select $1/4$ as our $r_d$.

\begin{table}[t!]
    \setlength\abovecaptionskip{0pt}
\setlength\belowcaptionskip{0pt}
    \caption{Studies on various activations in neural functions on NeurSTT on all six sequences from \citep{hui2019atr} in averaged $IoU$ and $F_1$ (\%).}
    \centering   
    {\scriptsize
    \begin{tabular}{p{2cm}<{\raggedright}p{1.2cm}<{\centering}p{1.2cm}<{\centering}}
    \hline
        Activation Function&$IoU$ $\uparrow$ & $F_1$ $\uparrow$\\\hline
        ReLU& 0.26 &0.65\\
        LeakyReLU& 0.36 &0.72 \\
        Tanh&7.33 &11.10\\
        \rowcolor{gray!20}Ours& \textbf{81.44}&\textbf{89.68} \\
        \hline
    \end{tabular}}
    \vspace{-0.3cm}
    \label{tab:activation}
\end{table}
\begin{table}[t!]
    \setlength\abovecaptionskip{0pt}
\setlength\belowcaptionskip{0pt}
    \caption{Studies on various background loss constraints in NeurSTT on all six sequences from \citep{hui2019atr} in averaged $IoU$ and $F_1$ (\%).}
    \centering   
    {\scriptsize
    \begin{tabular}{p{3cm}<{\raggedright}p{1cm}<{\centering}p{1cm}<{\centering}}
    \hline
        Loss Constraint for $\mathcal{B}^{\text{NLR}}$&$IoU$ $\uparrow$& $F_1$ $\uparrow$\\\hline
        $\mathcal{L}_{\text{MSE}} = \left\| \mathcal{D} - \mathcal{B}^{\text{NLR}} - \mathcal{T} \right\|_{F}^{2}  $& 75.96&85.77 \\
        \rowcolor{gray!20}$\mathcal{L}_{\text{Nuc}} = \sum\nolimits_{k = 1}^{n_3} {\left\| {\mathcal{B}_{k}^{\text{NLR}}} \right\|_{*}}$&\textbf{81.44} &\textbf{89.68}\\
        \hline
    \end{tabular}}
    \vspace{-0.3cm}
    \label{tab:loss}
\end{table}
\textbf{6) Impact of Activation Functions in NeurSTT:} We adopt the efficient and effective SIREN as our activation function to capture detailed and complex features. However, it is essential to study the impact of different activation functions on feature correlation for the ISTD task. We compare ReLU, LeakyReLU, and Tanh \citep{luo2023lrtfr} in Table \ref{tab:activation}. Notably, all compared methods are implemented implicitly, which is not as flexible as SIREN (as discussed in Section \ref{sec4}). Table \ref{tab:activation} shows that ReLU and LeakyReLU perform unsatisfactorily, while Tanh performs relatively better but still yields poor detection results. In comparison, SIREN proves more suitable for detection tasks, capturing complex features in spatial-temporal scenarios.


\begin{table}[t!]
    \vspace{-0.1cm}
\setlength\abovecaptionskip{0pt}
\setlength\belowcaptionskip{0pt}
    \caption{Ablation Studies on different key modules from NeurSTT on all six sequences from \citep{hui2019atr} in averaged $IoU$ and $F_1$ (\%).}
    \centering   
    {\scriptsize
    \begin{tabular}{p{0.8cm}<{\centering}p{1.5cm}<{\centering}p{1.5cm}<{\centering}p{1.1cm}<{\centering}p{1.1cm}<{\centering}}
    \hline
          Config. &Neur3DTV & Target &$IoU$ $\uparrow$&$F_1 $ $\uparrow$\\\hline
          I&\XSolidBrush  & \XSolidBrush & 0.26& 0.65\\
         {\textcolor{gray}{II}}&\XSolidBrush  & Substract & 28.68&40.81\\
         {\textcolor{SkyBlue}{III}}&\XSolidBrush  & $\mathcal{ST}$ & 80.15&88.91\\
         \rowcolor{gray!20}{\textcolor{Orchid}{IV}}&\Checkmark  & $\mathcal{ST}$ & \textbf{81.44}&\textbf{89.68}\\       
         V&NeurTV& $\mathcal{ST}$& 75.31&85.67\\
        VI&NeurSSTV& $\mathcal{ST}$&78.74 &87.89 \\
         VII&3DTV& $\mathcal{ST}$ &80.99 &89.38 \\
        \hline
    \end{tabular}}
    \label{tab:component}
        \vspace{-0.3cm}
\end{table}

\begin{figure}[t!]
\centering
\setlength\abovecaptionskip{-5pt}
\setlength\belowcaptionskip{-3pt}
  \includegraphics[width=0.6\linewidth]{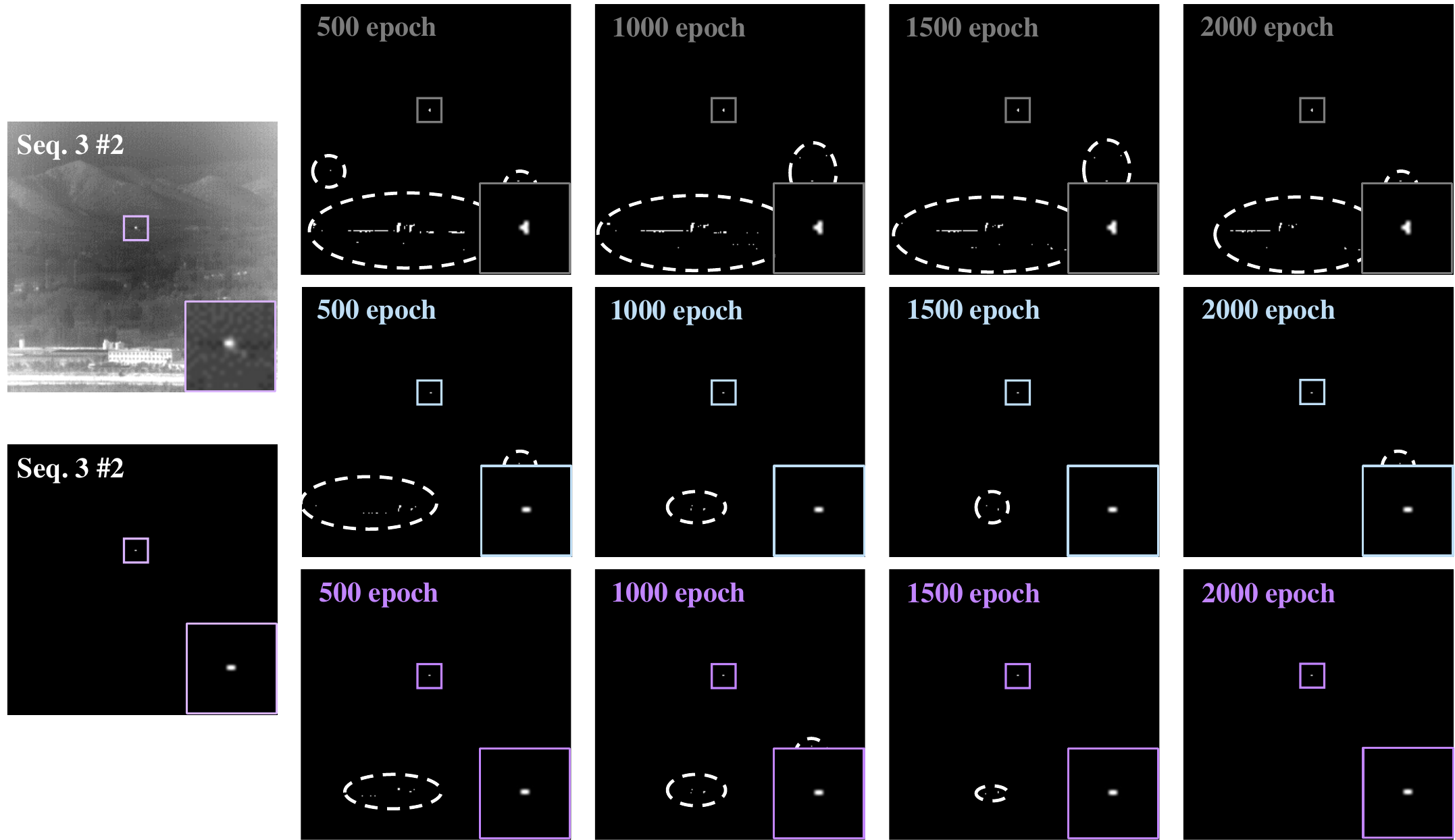}%
  \caption{Example of different configuration performance on different epochs on Seq. 3, where Config. II is in {\textcolor{gray}{gray}}, Config. III is in {\textcolor{SkyBlue}{skyblue}}, and Config. IV (NeurSTT) is in {\textcolor{Orchid}{purple}}. Notably, the proposed model can effectively reduce false alarms and preserve targets as the epoch increases.}
  \label{fig:epoch}
  \vspace{-0.3cm}
\end{figure}
\begin{figure}[t!]

\centering
\setlength\abovecaptionskip{-7pt}
\setlength\belowcaptionskip{-2pt}
  \includegraphics[width=0.8\linewidth]{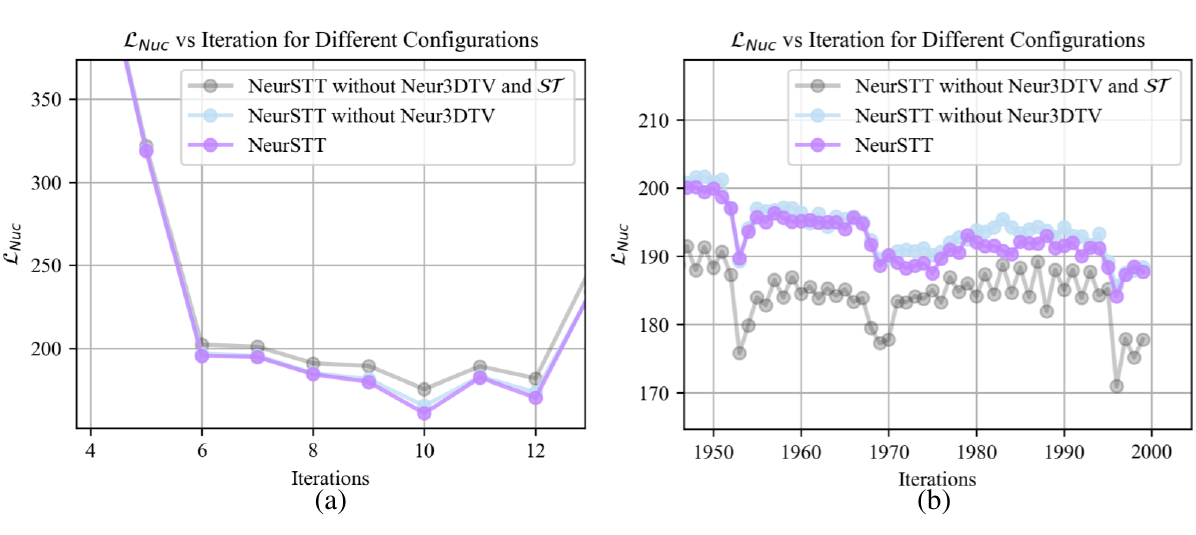}%
  \caption{Example of different nuclear norm loss $\mathcal{L}_{\text{Nuc}}$ performance via different configurations on Seq. 3. Our NeurSTT model is the fastest at recovering the background at early epochs while being relatively stable compared with the model without Neur3DTV.}
  \label{fig:loss}
  \vspace{-0.3cm}
\end{figure}

\textbf{7) Studies on NeurSTT's Key Modules:} In this section, we discuss the key modules of NeurSTT, including neural-represented low-rank background $\mathcal{B}^{\text{NLR}}$, neural regularizer Neur3DTV, and soft-thresholding operations ($\mathcal{ST}$) in target constraints. For the background, we examine the loss function used for regularization. Specifically, we compare our nuclear norm-based loss function $\mathcal{L}_{\text{Nuc}}$ with a general solution for handling HSI or MSI images, $\mathcal{L}_{\text{MSE}}$ (mean square error, MSE), in Table \ref{tab:loss}. Our results show that using $\mathcal{L}_{\text{Nuc}}$ outperforms $\mathcal{L}_{\text{MSE}}$ utilization by 5.48 in the $IoU$ (\%) metric, demonstrating that this constraint is more suitable for ISTD tasks.

For the target, we compared different combinations of target loss constraints and the Neur3DTV involvement in Table \ref{tab:component}. Here, {\small \XSolidBrush} means no such module, while {\small  \Checkmark} means it does; "Subtract"  means the loss constrain: $\|\mathcal{D}-\mathcal{B}^{\text{NLR}} \|_{1}$, without soft-thresholding, as \citep{zhang20233dstpm}; and $\mathcal{ST}$ means our method in Eq. (\ref{eq_lossT}). We observe that introducing the $l_1$-norm as the target constraint element is effective when comparing Config. I and II. Config. III shows that using a general $\mathcal{ST}$ operator significantly improves detection performance by focusing more on targets rather than just background recovery. The result of Config. IV shows that incorporating background regularization through Neur3DTV improves detection performance by 1.28 in the $IoU$ (\%) metric compared to Config. III.

For comparison, we visualize results at different epochs with various configurations: Config. II in {\textcolor{gray}{II}}, Config. III in {\textcolor{SkyBlue}{blue}}, and Config. IV (our setting) in {\textcolor{Orchid}{purple}}. As shown in Fig. \ref{fig:epoch}, numerous false alarms occur when only subtracting the original and background. This issue is significantly mitigated by leveraging $\mathcal{ST}$, even in early epochs. In comparison, our Neur3DTV-enhanced model exhibits fewer clusters in the initial stages and reduces all false alarms when reaching the maximum iterations.

Moreover, $\mathcal{L}_{\text{Nuc}}$ is set as the background loss function, serving as both a background regularizer and an index for background recovery. As shown in Fig. \ref{fig:loss}, due to the low-rank nature of infrared images, both methods quickly reach low loss values early on. Notably, $\mathcal{ST}$-embedded methods achieve the lowest loss values, while Config. II performs slightly less, demonstrating the effectiveness of traditional operator introduction. By the end of the iterations, all three methods recover the background well. Although Config. II shows slightly better recovery performance, it focuses too much on the background, resulting in poor target performance, as depicted in Table \ref{tab:component}. In contrast, our NeurSTT balances both tasks effectively, with its loss being lower than that of Config. II and III. The ablation studies demonstrate the contribution of each module in our NeurSTT model.

We also study different implementations of neural total variation in Table \ref{tab:component}, denoted as Configs. V to VII, including: NeurTV, NeurSSTV \citep{luo2024neurtv}, and traditional 3DTV (as discussed in Section \ref{sec:4.3}). Compared with Config. III, the introduction of NeurTV and NeurSSTV shows side effects as the performance index decreases, indicating that they are not suitable for detection tasks. As shown in the first row of Config. IV in Table \ref{tab:parameter}, the slight introduction of temporal information aids the detection task compared to using only spatial information in NeurTV. In contrast, differential-based 3DTV slightly improves detection performance and is more suitable for this task. Our NeurSTT combines the advantages of both approaches, resulting in better detection performance.
\begin{figure*}[t!]
\centering
\setlength\abovecaptionskip{-5pt}
\setlength\belowcaptionskip{-1pt}
  \includegraphics[width=0.95\linewidth]{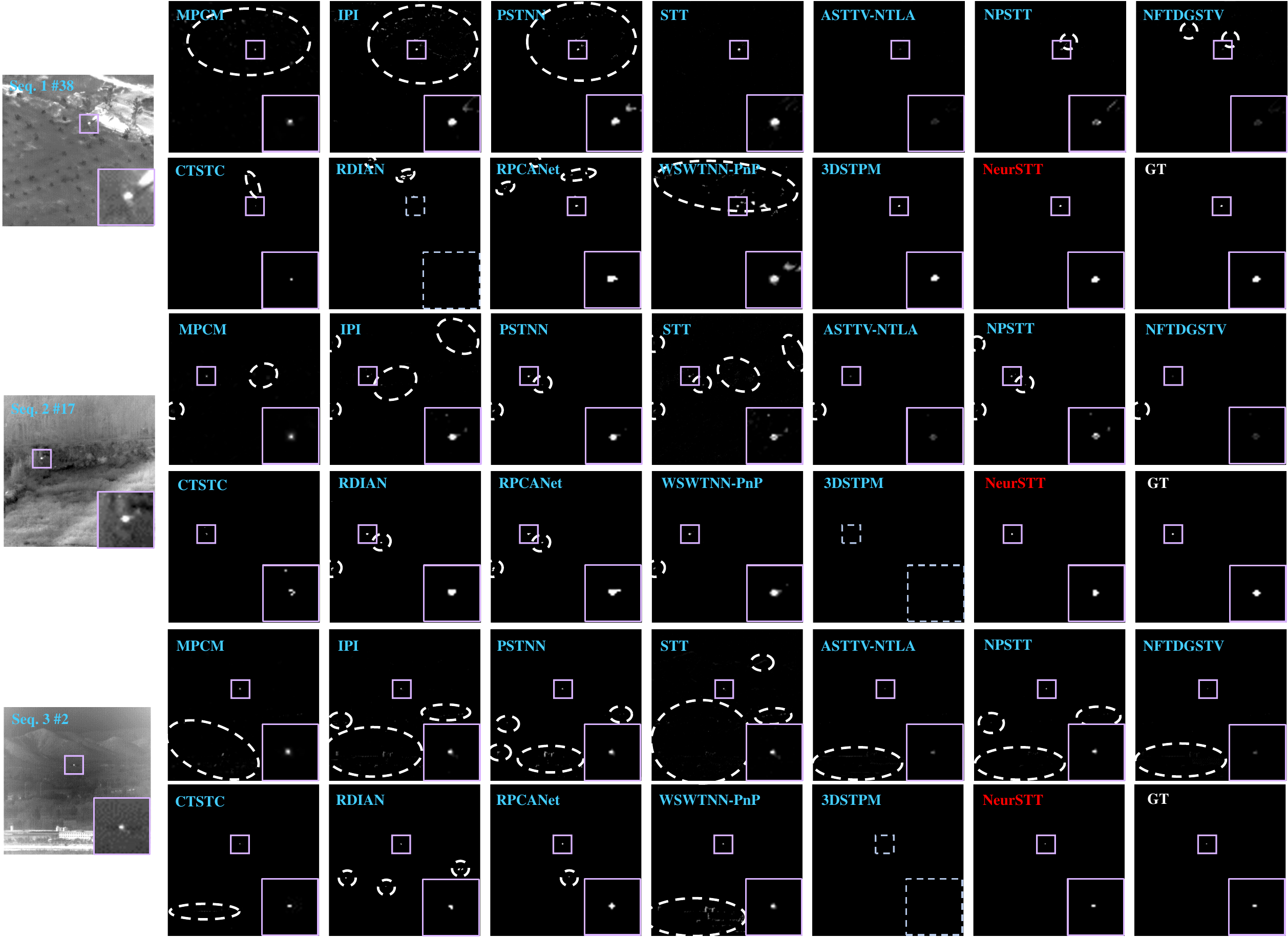}%
  \caption{Visual comparison of detection results from different algorithms and labeled ground truth on Seqs. 1 to 3. {\textcolor{CarnationPink}{Pink}} boxes indicate the correct detection, white dot circles mean the false alarms, and {\textcolor{SkyBlue}{skyblue}} dot boxes stand for the miss detection.}
  \label{fig:visual1-3}
        \vspace{-0.3cm}
\end{figure*}
\begin{figure*}[t!]
\centering
\setlength\abovecaptionskip{-5pt}
\setlength\belowcaptionskip{-1pt}
  \includegraphics[width=0.95\linewidth]{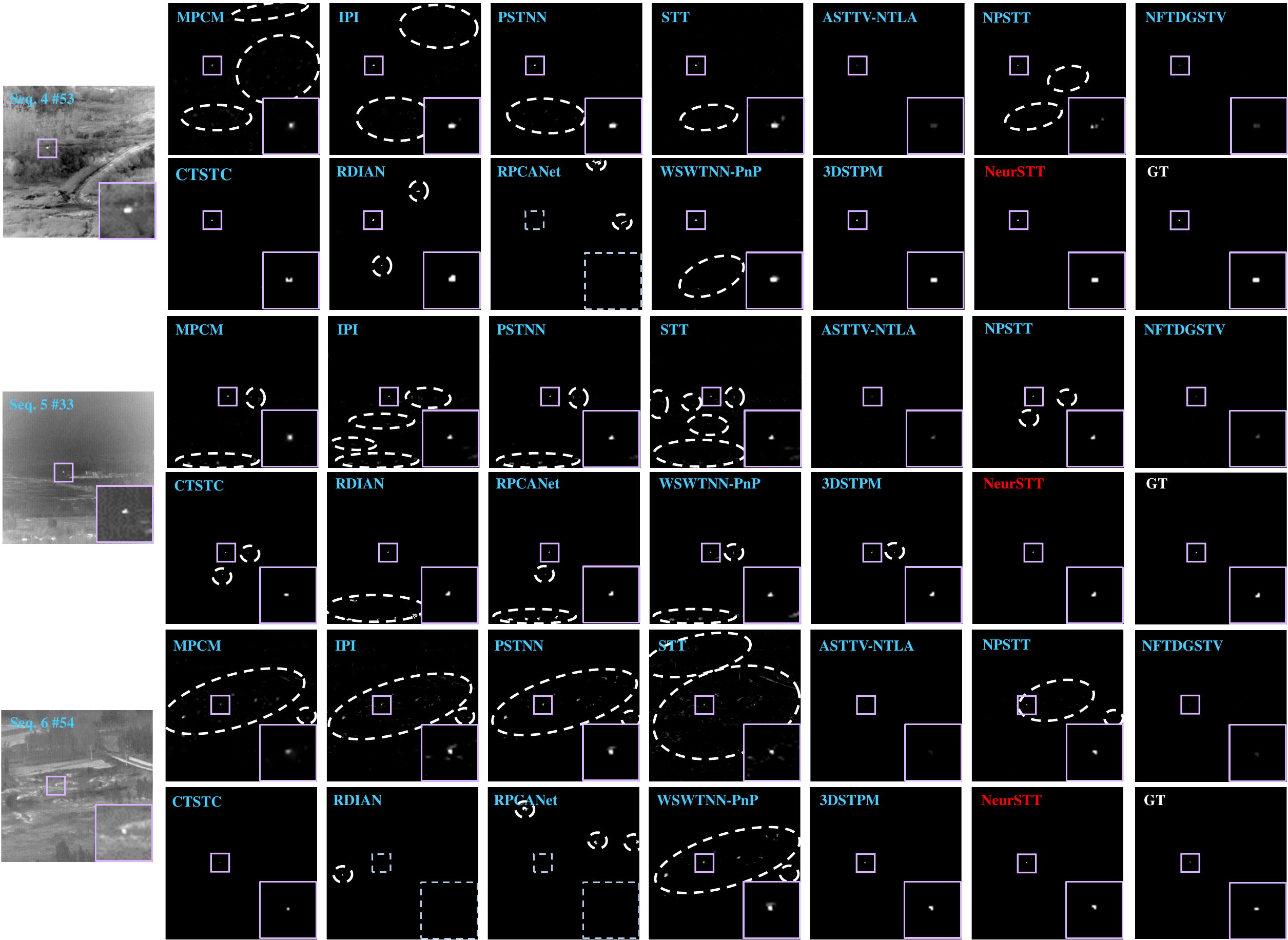}%
  \caption{Visual comparison of detection results from different algorithms and labeled ground truth on Seqs. 4 to 6. {\textcolor{CarnationPink}{Pink}} boxes indicate the correct detection, white dot circles mean the false alarms, and {\textcolor{SkyBlue}{skyblue}} dot boxes stand for the miss detection.}
    \label{fig:visual4-6}
      \vspace{-0.3cm}
\end{figure*}

\subsection{Comparison with Baselines}
In this section, we compare our NeurSTT with SOTA methods across nine sequences to demonstrate its effectiveness, both visually and numerically.

\begin{figure*}[t!]
\setlength\abovecaptionskip{-5pt}
\setlength\belowcaptionskip{-1pt}
\centering
  \includegraphics[width=0.95\linewidth]{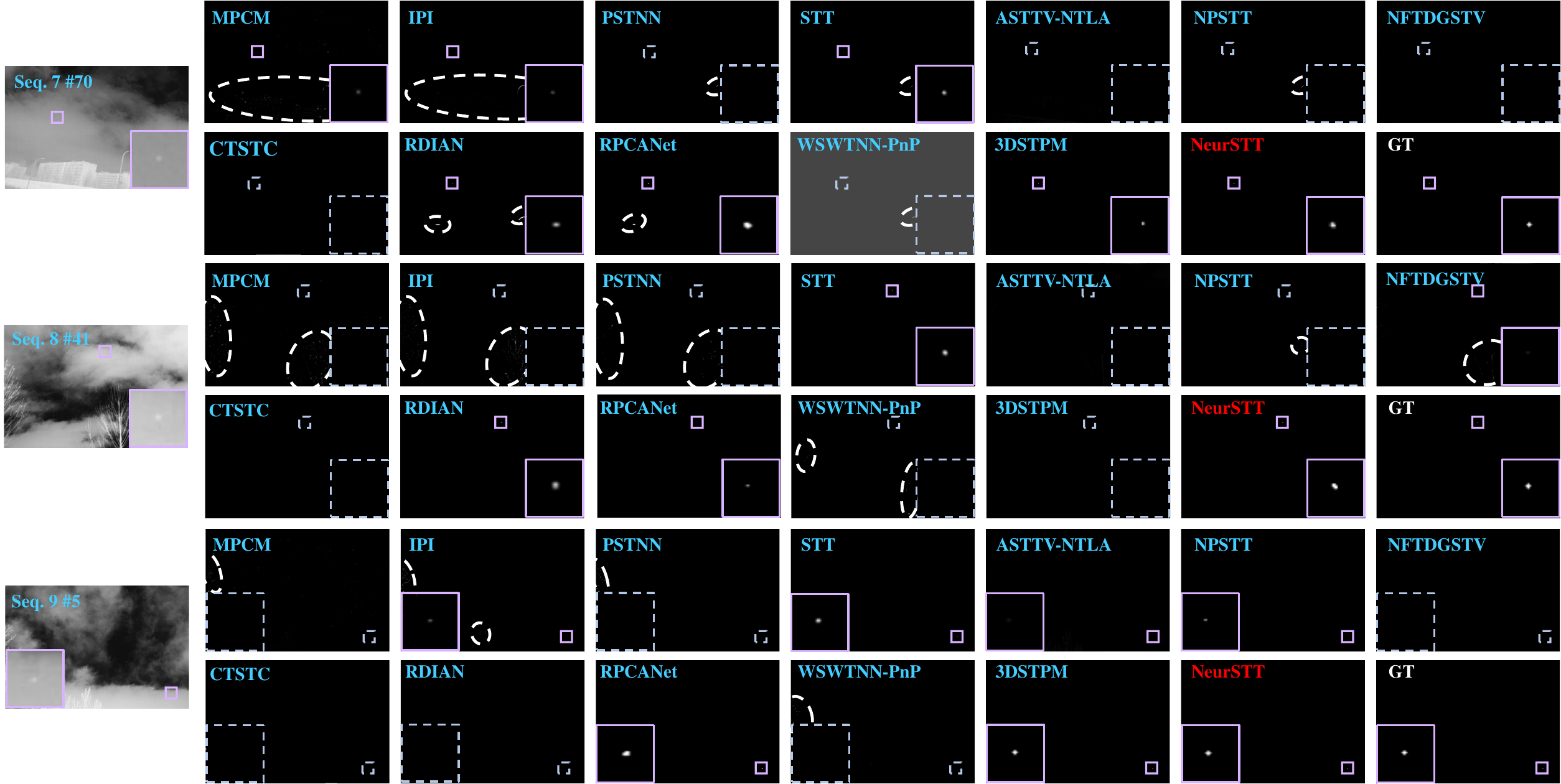}%
  \caption{Visual comparison of detection results from different algorithms and labeled ground truth on Seqs. 7 to 9. {\textcolor{CarnationPink}{Pink}} boxes indicate the correct detection, white dot circles mean the false alarms, and {\textcolor{SkyBlue}{skyblue}} dot boxes stand for the miss detection.}
    \label{fig:visual7-9}
          \vspace{-0.2cm}
\end{figure*}

\textbf{1) Visual Comparison:} We present representative frames from different sequences in Figs. \ref{fig:visual1-3}, \ref{fig:visual4-6}, and \ref{fig:visual7-9}. For medium-sized sequences, single-frame methods like MPCM, IPI, and PSTNN detect targets but generate many false alarms. Deep learning methods such as RDIAN and RPCANet struggle to distinguish real targets from target-like ones, resulting in false alarms, as seen in Seq. 3 from RDIAN. Notably, RPCANet misses detections in Seqs. 4 and 6, incorrectly recognizing other clusters as real targets.

In comparison, multi-frame-based approaches reduce false alarms while preserving targets when they occupy more pixels. However, many of these methods suffer from the "dead point" issue, where parts of the camera are broken, generating moving points, as observed in Seq. 3. Unsupervised learning-based methods like 3DSTPM and our NeurSTT produce fewer false alarms due to the utilization of sequential information and the nonlinearity of neural networks. Compared to the miss-detections in Seqs. 2 and 3 from 3DSTPM, our NeurSTT demonstrates better robustness and detects all targets in the representative frames.

However, when scenes involve larger sequences, as shown in Fig. \ref{fig:visual7-9}, both single and multi-frame-based algorithms encounter significant miss-detection issues. Methods like PSTNN, CTSTC, and WSWTNN-PnP fail in all three sequences. In comparison, the spatial-temporal patch tensor-based STT can detect almost all targets, although there are still several false alarms in Seq. 7. The deep learning method RPCANet can almost detect the tiny target when it has a relatively high target-to-background contrast, but RDIAN performs poorly, detecting targets with false alarms in Seq. 7.

Our NeurSTT, on the other hand, successfully detects all the targets in all three sequences, whereas its competitor 3DSTPM does not. NeurSTT also maintains similar target sizes compared to the ground truth labels. In summary, we can conclude that the proposed NeurSTT outperforms the compared ISTD baselines in visual measurement.

\textbf{2) Numerical Comparison:} To highlight the effectiveness of our NeurSTT over other SOTA methods, we adopt target-level and pixel-level metrics. The symbols $\uparrow$ and $\downarrow$ denote metrics where higher and lower values are preferable, respectively.

For 3-D ROC metrics, we present the averaged AUC values from nine sequences in Table \ref{tab:3droc} and showcase their performance on 3-ROC in Figs. \ref{fig:3droc1}, \ref{fig:3droc2}, and \ref{fig:3droc3}. In Table \ref{tab:3droc}, the best metric is bolded, and the second-best is underlined. The numerical metrics align well with the visual comparison. For instance, approaches like STT, ASTTV-NTLA, and 3DSTPM achieve satisfactory results in $\text{AUC}_{(\text{FPR},\text{TPR})}$. However, some single-frame methods, despite having relatively high $\text{AUC}_{(\text{FPR},\text{TPR})}$, fail to address the high number of false alarms, resulting in poor overall performance (e.g., MPCM has 0.9656 for $\text{AUC}_{(\text{FPR},\text{TPR})}$ but 0.3988 for $\text{AUC}_{\text{TDBS}}$).

\begin{table*}[t!]
\setlength{\abovecaptionskip}{0pt}
\setlength{\belowcaptionskip}{2pt}
\caption{Comparison of 3-D ROC's AUC metrics on the averaged value on different sequences with other SOTA methods.}
\label{tab:3droc}
\centering
\resizebox{\linewidth}{!}{
\begin{tabular}{ccccccccc}
\hline
 Method & AUC$_{(\text{FPR},\text{TPR})}$ $\uparrow$& AUC$_{(\tau,\text{FPR})}$ $\downarrow$ & AUC$_{(\tau,\text{TPR})}$  $\uparrow$ &$\text{AUC}{_{\text{BS}}}$ $\uparrow$&$\text{AUC}{_{\text{TD}}}$ $\uparrow$ &$\text{AUC}{_{\text{TDBS}}}$ $\uparrow$& $\text{AUC}_{\text{ODP}}$ $\uparrow$& $\text{AUC}{_{\text{SNPR}}}$ $\uparrow$ \\
\midrule
        MPCM \citep{wei2016mpcm}& 0.9656  & 0.0065  & 0.4054  & 0.9591  & 1.3710  & 0.3988  & 1.3645  & 64.0874  \\
         IPI \citep{gao2013ipi}&0.9660  & 0.0062  & 0.6513  & 0.9598  & 1.6173  & 0.6451  & 1.6111  & 105.5405  \\
        PSTNN \citep{zhang2019pstnn}& 0.8338  & 0.0055  & 0.6070  & 0.8283  & 1.4408  & 0.6015  & 1.4353  & 108.6086  \\
       STT \citep{liu2020small}& \underline{0.9832}  & 0.0061  & \underline{0.7581}  & \underline{0.9772}  & \underline{1.7413}  & \underline{0.7520}  & \underline{1.7352}  & 127.1572  \\ 
        ASTTV-NTLA \citep{liu2021asttv}&0.9225  & 0.0063  & 0.5803  & 0.9162  & 1.5029  & 0.5740  & 1.4966  & 106.8721  \\ 
        NPSTT \citep{wang2021npstt}&0.8045  & \underline{0.0051}  & 0.4296  & 0.7994  & 1.2342  & 0.4246  & 1.2291  & 84.2126  \\
        NFTDGSTV \citep{liu2023tucker}&0.8613  & 0.0106  & 0.5009  & 0.8507  & 1.3622  & 0.4902  & 1.3516  & 81.0618  \\
        CTSTC \citep{luo2024clustering}&0.7309  & 0.0052  & 0.4201  & 0.7257  & 1.1509  & 0.4292  & 1.1682  & 85.4648  \\ 
        RDIAN \citep{sun2023rdian}&0.8142  & 0.0056  & 0.6332  & 0.8222  & 1.4609  & 0.6244  & 1.4554  & 110.5442  \\
        RPCANet \citep{wu2024rpcanet}&0.8118  & 0.0054  & 0.6085  & 0.8063  & 1.4202  & 0.6030  & 1.4148  & 109.2852  \\
        WSWTNN-PnP \citep{liu2023wswtnn-pnp}&0.8319  & 0.0481  & 0.6506  & 0.7838  & 1.4825  & 0.6025  & 1.4344  & 77.8793  \\ 
        3DSTPM \citep{zhang20233dstpm}& 0.8656  & \textbf{0.0050}  & 0.7253  & 0.8606  & 1.5910  & 0.7203  & 1.5860  & \underline{144.3647}  \\ 
        \rowcolor{gray!20}NeurSTT &\textbf{0.9849}  & \textbf{0.0050}  & \textbf{0.9599}  & \textbf{0.9797}  & \textbf{1.9447}  & \textbf{0.9549}  & \textbf{1.9397}  & \textbf{191.6919} \\
\hline
\end{tabular}}
\vspace{-0.2cm}
\end{table*}
\begin{figure*}[t!]
\vspace{-0.2cm}
\centering
\setlength\abovecaptionskip{-3pt}
\setlength\belowcaptionskip{0pt}
  \includegraphics[width=0.95\linewidth]{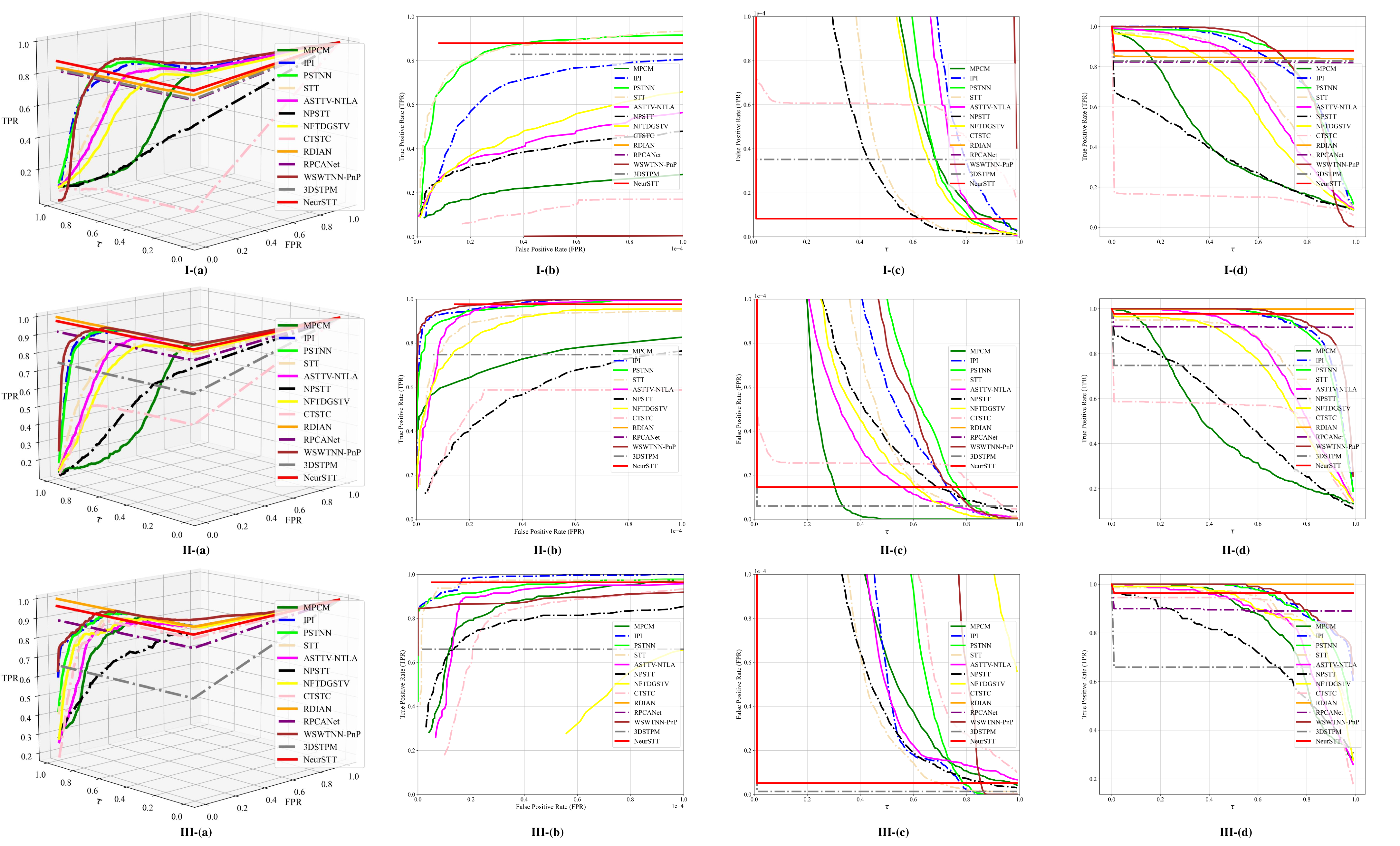}%
  \caption{3-D ROC curves and corresponding 2-D ROC curves (ROC$_{(\text{FPR},\text{TPR})}$, ROC$_{(\tau,\text{FPR})}$, and ROC$_{(\tau,\text{TPR})}$) of various methods on Seqs. 1 to 3.}
  \label{fig:3droc1}
  \vspace{-0.3cm}
\end{figure*}
\begin{figure*}[t!]
\vspace{-0.1cm}
\centering
\setlength\abovecaptionskip{-3pt}
\setlength\belowcaptionskip{0pt}
  \includegraphics[width=0.95\linewidth]{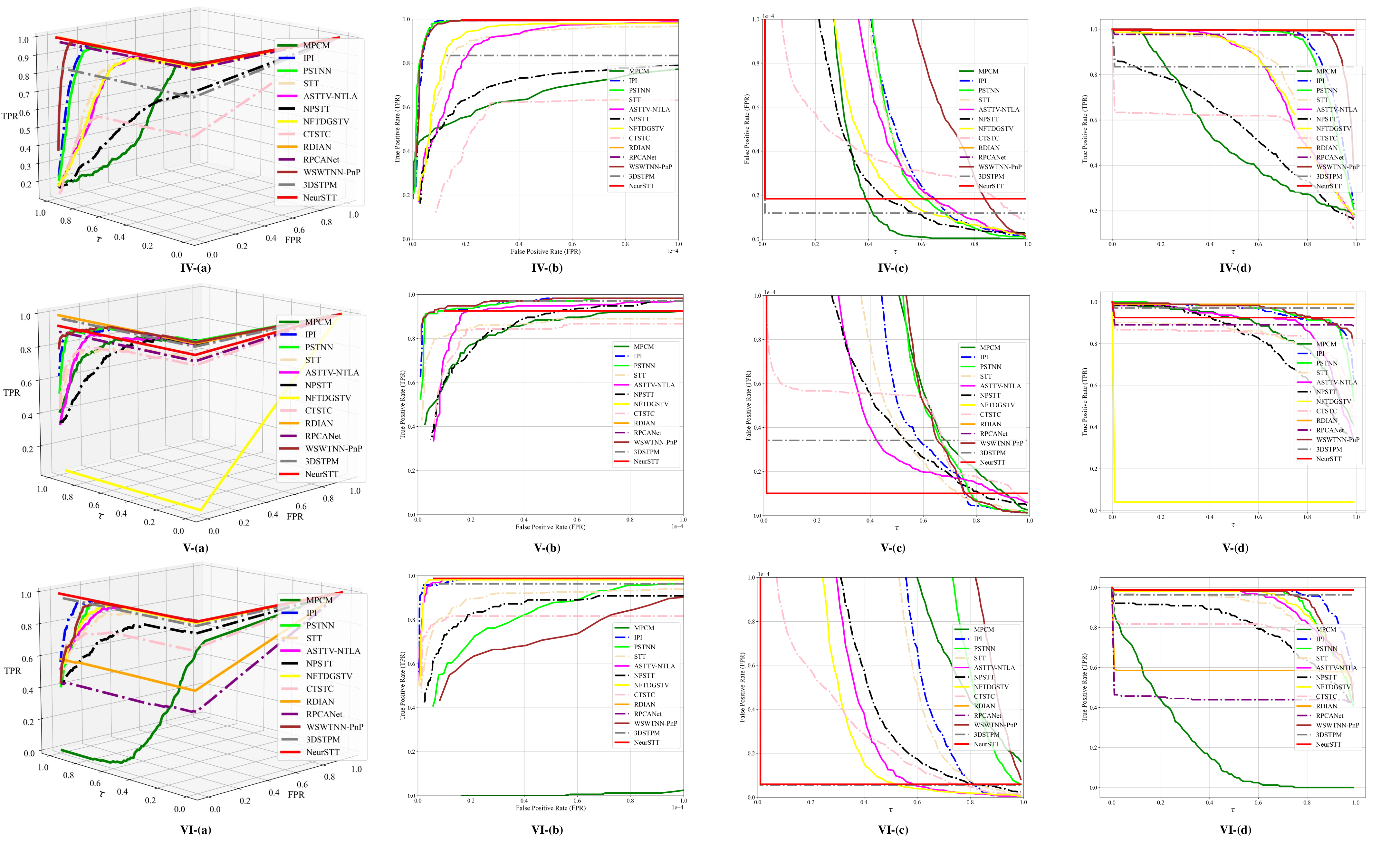}%
  \caption{3-D ROC curves and corresponding 2-D ROC curves (ROC$_{(\text{FPR},\text{TPR})}$, ROC$_{(\tau,\text{FPR})}$, and ROC$_{(\tau,\text{TPR})}$) of various methods on Seqs. 4 to 6.}
    \label{fig:3droc2}
    \vspace{-0.3cm}
\end{figure*}
\begin{figure*}[t!]
\centering
\vspace{-0.1cm}
\setlength\abovecaptionskip{-3pt}
\setlength\belowcaptionskip{0pt}
  \includegraphics[width=0.95\linewidth]{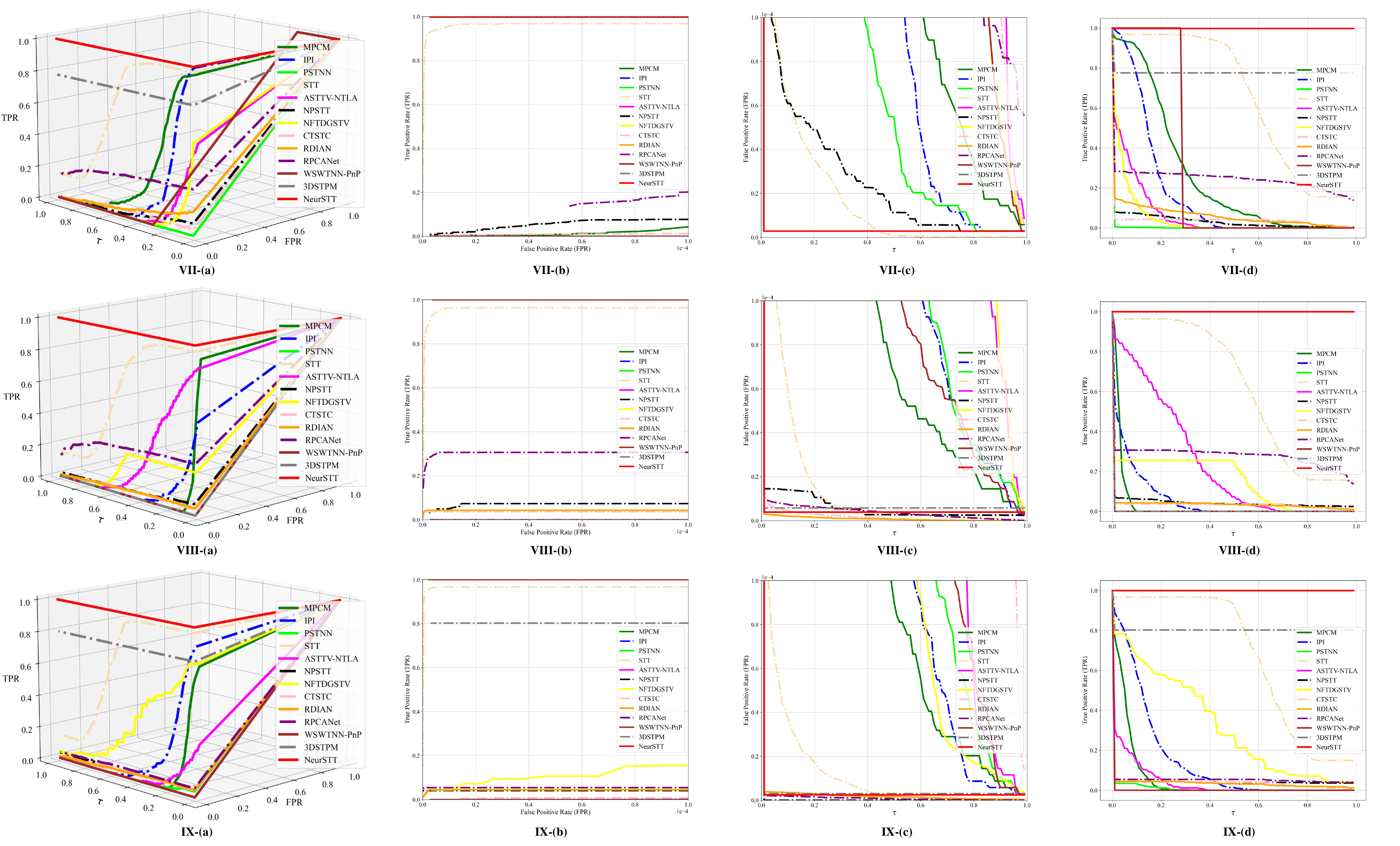}%
  \caption{3-D ROC curves and corresponding 2-D ROC curves (ROC$_{(\text{FPR},\text{TPR})}$, ROC$_{(\tau,\text{FPR})}$, and ROC$_{(\tau,\text{TPR})}$) of various methods on Seqs. 7 to 9.}
    \label{fig:3droc3}
    \vspace{-0.3cm}
\end{figure*}
In contrast, algorithms that consider temporal information better suppress false alarms. For example, NPSTT and CTSTC have lower $\text{AUC}_{(\tau,\text{FPR})}$ values, 0.0051 and 0.0052, respectively. However, due to their poor performance in Seqs. 7 to 9, they have relatively low $\text{AUC}_{\text{ODP}}$ values of 1.2291 and 1.1682, respectively. STT, leveraging both local and global spatial-temporal information, shows almost suboptimal performance except for $\text{AUC}_{(\tau,\text{FPR})}$ and $\text{AUC}_{\text{SNPR}}$. 
The proposed NeurSTT surpasses the compared methods in all AUC measurements, with a notable 32.42\% improvement over the suboptimal 3DSTPM in $\text{AUC}_{\text{SNPR}}$, demonstrating its superiority in both target detectability and background suppression.

We present the 3-D ROC curves and corresponding 2-D ROC plots in Figs. \ref{fig:3droc1}, \ref{fig:3droc2}, and \ref{fig:3droc3}. For clarity, we zoom in on the FPR axis in $\text{ROC}_{(\text{FPR},\text{TPR})}$ and the TPR axis in $\text{AUC}_{(\tau,\text{TPR})}$ to $\times 10^{-4}$. In Figs. \ref{fig:3droc1} and \ref{fig:3droc2}, methods like CTSTC, NFTDGSTV, and 3DSTPM perform well on specific sequences but occasionally fail (e.g., CTSTC in Seqs. 1, 2, and 4; NFTDGSTV in Seq. 5; 3DSTPM in Seq. 3). Deep learning methods such as RPCANet and RDIAN are effective except for Seq. 6, where they suffer miss-detections due to target-like ground interferences. In comparison, our NeurSTT quickly reaches the top-center corner in almost all 3-D ROC curves, demonstrating excellent background suppression ability, as shown in I-(c) and V-(c) in Figs. \ref{fig:3droc1} and \ref{fig:3droc2}.

\begin{table*}[t!]
\setlength{\abovecaptionskip}{0pt}
\setlength{\belowcaptionskip}{0pt}
\caption{Numerical $IoU$ and $F_1$ (\%) comparison of different SOTA methods and NeurSTT on Seqs. 1 to 9, and the average values.}
\centering
\resizebox{\linewidth}{!}{
\begin{tabular}{ccccccccccccccccccccc}
\hline
\multirow{2}{*}{Method} &\multicolumn{2}{c}{Seq.1}  &\multicolumn{2}{c}{Seq.2} & \multicolumn{2}{c}{Seq.3} & \multicolumn{2}{c}{Seq.4} & \multicolumn{2}{c}{Seq.5} & \multicolumn{2}{c}{Seq.6}& \multicolumn{2}{c}{Seq.7} & \multicolumn{2}{c}{Seq.8}& \multicolumn{2}{c}{Seq.9}& \multicolumn{2}{c}{Avg} \\\cline{2-21}
& $IoU$& $F_1$& $IoU$&$F_1$ & $IoU$&$F_1$ & $IoU$& $F_1$& $IoU$& $F_1$& $IoU$&$F_1$ & $IoU$& $F_1$&$IoU$ &$F_1$&$IoU$&$F_1$&$IoU$&$F_1$\\
\midrule
 MPCM \citep{wei2016mpcm}&16.32 &28.06 &38.29 & 55.38& 37.43&54.47 &43.37 &60.50 &21.80 &35.79 & 0.93&1.85 & 0.34&0.68 &0.00 &0.00&0.10&0.20&17.62&26.33\\
 IPI \citep{gao2013ipi}&27.25 &42.82 & 65.39&79.07 & 43.11& 60.25& 61.30& 76.00& 38.03&55.11 &14.94 &26.00 &0.83 & 1.65&0.05 &0.10&0.28&0.56&27.91&37.95\\
PSTNN \citep{zhang2019pstnn}& 39.09&56.20 & 52.97& 69.26&16.43 &28.22 & 65.74& 79.33& 18.40& 31.08& 6.68&12.53 &0.02 &0.03 & 0.00&0.00&0.03&0.07&22.15&30.75\\
STT \citep{liu2020small}& \underline{72.26}& \underline{83.90}& 69.13&81.74 & 58.52&73.83 & 61.06& 75.82& 38.66& 55.76& 17.73& 30.12&\underline{21.34} &\underline{35.17} &\underline{22.33} &\underline{36.51}&32.47&49.02&43.72&57.99\\
ASTTV-NTLA \citep{liu2021asttv}& 25.25& 40.32& \underline{79.32}& \underline{88.46}& 43.76& 60.88& 63.33&77.55 &\underline{52.27} &\underline{68.66} &70.74 & 82.86& 0.06&0.12 & 0.16&0.31&0.02&0.03&37.21&46.58\\
NPSTT \citep{wang2021npstt}&29.19 &45.19 &41.96 &59.12 & 42.86& 60.00&50.10 &66.75 & 40.69& 57.85& 44.90& 61.98&1.65 &3.25 & 3.53&6.82&0.00&0.00&28.32&40.11\\
NFTDGSTV \citep{liu2023tucker}& 39.11&56.23 & 69.22& 81.81& 3.33& 6.44& \underline{75.37}&\underline{85.96} &0.06 & 0.09&80.50 &89.19 &0.06 & 0.13& 0.14&0.28&0.30&0.60&29.79&35.64\\
CTSTC \citep{luo2024clustering}&11.15 &20.06 &46.92 &63.87 & 23.66&38.26 & 44.02& 61.13& 32.17&48.68& 0.10&0.21 & 0.04& 0.10&0.00 &0.00&0.13&0.26&17.58&25.84\\
RDIAN \citep{sun2023rdian}&10.36 &18.77 & 22.59& 36.85& 2.29& 4.47& 16.72& 28.65& 1.96& 3.85& 3.03&5.89 & 0.21& 0.41& 0.00&0.00&3.63&7.00&6.75&11.77\\
RPCANet \citep{wu2024rpcanet}&16.52 &28.36 & 31.04&47.38 & 5.98& 11.29& 9.54&17.42& 2.14& 4.19& 2.68&5.22 & 2.81& 5.47&19.36 &32.44&4.85&9.26&10.55&17.89\\
WSWTNN-PnP \citep{liu2023wswtnn-pnp}&7.55 &14.03 & 59.10& 74.29& 3.62&6.98 &35.76 &52.68 & 16.83&28.81 &4.84 & 9.24&0.00 & 0.00& 0.00&0.00&0.00&0.00&14.19&20.67\\
3DSTPM \citep{zhang20233dstpm}&68.60&81.37&71.07&83.09 &\underline{67.40}&\underline{80.52}&72.86&84.30&47.73&64.62&\underline{82.29}&\underline{90.29}&11.62&20.82&0.00&0.00&\underline{69.12}&\underline{81.74}&\underline{54.52}&\underline{65.19}\\
\rowcolor{gray!20}NeurSTT&\textbf{80.97}&\textbf{89.49}&\textbf{86.65}&\textbf{92.86}&\textbf{85.83}&\textbf{92.37}&\textbf{81.30}&\textbf{89.68}&\textbf{70.80}&\textbf{82.90}&\textbf{83.08}&\textbf{90.76}&\textbf{86.21}&\textbf{92.60}&\textbf{82.03}&\textbf{90.13}&\textbf{88.51}&\textbf{93.90}&\textbf{82.82}&\textbf{90.52}\\
\hline
\end{tabular}}
    \vspace{-0.2cm}
\label{tab:pixel}
\end{table*}
For Seqs. 7 to 9, our model’s effectiveness is more distinct compared to other methods, as shown in Fig. \ref{fig:3droc3}. Except for STT, most traditional methods fail in these sequences due to poor correlation of local and global features. The unsupervised learning-based 3DSTPM completes detection in Seqs. 7 and 9 but fails in Seq. 8. Our proposed architecture succeeds in all sequences by considering both temporal and spatial features, with neural layers effectively correlating local and global information.

To examine pixel detection performance, we present the $IoU$ and $F_1$ in Table \ref{tab:pixel}. The best values are in bold, and sub-optimal ones are underlined. From Table \ref{tab:pixel}, our method achieves the highest measurements in both metrics across all sequences and averages. We can conclude that NeurSTT outperforms others from the following points: 

Optimization-based methods are superior to other strategies. For instance, matrix and tensor-based single-frame methods IPI and PSTNN lead in $IoU$ by 58.40\% and 25.71\%, compared with MPCM, respectively. Additionally, the optimization-inspired RPCANet outperforms the "black box" RDIAN by 52.00\% in $F_1$. 

In addition, temporal-based methods outperform single-frame methods. For example, as an almost temporal extension of PSTNN, STT exceeds single-frame methods by an average of 21.57 in $IoU$ (\%). Specifically, PSTNN experiences massive miss-detections in Seqs. 3, 7, and 8 (fails to detect anything), while STT is the second-best algorithm in Seqs. 7 and 8, and the third-best in Seq. 3. Sequential-inspired methods such as ASTTV-NTLA and NFTDGSTV also emerge as sub-optimal approaches.

Moreover, nonlinear neural layers have a better capability of feature representation than traditional linear operators. Although WSWTNN-PnP introduces a PnP denoiser in optimization processes, it suffers from miss-detections and false alarms due to using a denoise model designed for visible light, which differs significantly from infrared imagery. Additionally, there is a risk that targets are mistaken for "noise" as image size increases, leading to failure in Seqs. 7 to 9. Therefore, it is crucial to balance learning and optimization models in the sequential domain.

As depicted in the last two columns of Table \ref{tab:pixel}, neural-based methods 3DSTPM and our NeurSTT achieve the second-best and best average measurements, respectively. Compared to the best traditional spatial-temporal method, STT, they have leads of 24.70\% and 89.43\% in $IoU$, respectively. However, as 3DSTPM experiences miss-detections in Seq. 8, our NeurSTT demonstrates better robustness in both $IoU$ and $F_1$. Unlike 3DSTPM, which primarily focuses on background estimation, NeurSTT emphasizes target regularization and the correlation of the continuous temporal domain. Therefore, it is reasonable to conclude that our NeurSTT is more effective and appropriate for ISTD tasks.

\begin{table*}[]
\setlength{\abovecaptionskip}{0pt}
\setlength{\belowcaptionskip}{0pt}
\caption{Running time comparison of different SOTA methods on Seqs.1 to 9, and the average values. (\textit{Optim.} is short for optimization, \textit{Multi.} is short for multi-frame.)}
\centering
\resizebox{\linewidth}{!}{
{\scriptsize
\begin{tabular}{ccccccccccccc}
\hline
Method& Category&Scene&Seq.1  &Seq.2 &Seq.3 &Seq.4 &Seq.5& Seq.6& Seq.7 &Seq.8& Seq.9& Avg \\
\midrule
        MPCM \citep{wei2016mpcm}& \textit{HVS}&\textit{Single}& 0.039 &0.037  &0.039  &0.038  & 0.039&0.037& 0.037  &  0.241& 0.240& 0.083 \\
         IPI \citep{gao2013ipi}&\textit{Optim.}&\textit{Single}&5.258 & 5.197 &5.538  &4.901  &5.799 &5.167&33.477   & 33.968 & 35.437&14.971 \\
        PSTNN \citep{zhang2019pstnn}&\textit{Optim.}&\textit{Single}&0.253  &0.118  & 0.250 & 0.193 &0.214 &0.274   & 0.934 &0.931 &1.021&0.465 \\\hline
       STT \citep{liu2020small}&\textit{Optim.}&\textit{Multi.}&3.792 &3.791  &4.140  &3.682 &3.899 &3.639 &  23.621 & 23.525& 29.706& 11.088 \\ 
        ASTTV-NTLA \citep{liu2021asttv}&\textit{Optim.}&\textit{Multi.}& 2.979 & 2.811 &2.839 &2.895&2.829  &2.862 & 14.529 & 14.591 & 14.529& 6.763 \\ 
        NPSTT \citep{wang2021npstt}&\textit{Optim.}&\textit{Multi.}& 5.259 &2.205  &3.198  & 4.086&4.404 & 5.450  &17.226  &17.855 &18.837& 8.724\\
        NFTDGSTV \citep{liu2023tucker}&\textit{Optim.}&\textit{Multi.}& 2.200 & 2.288 &2.150  & 2.325 & 2.276& 2.267  & 12.177 & 12.85&13.793&5.814  \\
        CTSTC \citep{luo2024clustering}&\textit{Optim.}&\textit{Multi.}& 0.206 & 0.183 &0.217  &0.186  &0.169 &0.166   & 0.657 &1.278 &0.721& 0.420\\ \hline
        RDIAN \citep{sun2023rdian}&\textit{DL}&\textit{Single}& 0.006 & 0.005 & 0.006 & 0.006 &0.006 &0.005   &  0.018& 0.019&0.019& 0.010 \\
        RPCANet \citep{wu2024rpcanet}&\textit{DL}&\textit{Single}& 0.002 & 0.003 & 0.002 & 0.003 & 0.002&0.002   & 0.008 &0.007 &0.008& 0.004 \\
        WSWTNN-PnP \citep{liu2023wswtnn-pnp}&\textit{DL}&\textit{Multi.}&  2.582& 1.088 & 2.913 &1.532  &1.938 & 2.368  & 14.388 &9.609 &11.250& 5.296\\ 
        3DSTPM \citep{zhang20233dstpm}&\textit{DL}&\textit{Multi.}&1.726 & 1.713 & 1.650 & 1.726&1.695 &1.665   &6.750  & 6.725&6.750&3.378  \\ 
        \rowcolor{gray!20}NeurSTT&\textit{DL}&\textit{Multi.}&  0.809& 0.885 &0.878  &0.878  &0.869 & 0.882  & 1.806 & 1.859&1.828&1.188 \\
\hline
\end{tabular}}}
\label{tab:time}
\vspace{-0.2cm}
\end{table*}

\textbf{3) Computational Comparison:} Computational performance is another essential metric in ISTD algorithm evaluations. We present the running times of all tested methods on different sequences in Table \ref{tab:time}. On average, multi-frame methods have higher running times than single-frame methods due to larger video inputs and time-consuming tensor decomposition operations. Moreover, as image size increases, running times rise dramatically; for instance, STT takes 29.706 seconds to process a $720\times480$ image on average. Single-frame DL methods are among the fastest since they use pre-trained weights on a GPU without considering temporal information. In contrast, multi-frame DL-based WSWTNN-PnP, despite GPU acceleration, still faces extensive tensor decomposition calculations.

In comparison, 3DSTPM methods are relatively faster since they do not include traditional operators in their networks. However, due to the use of 3DCNNs, their execution times and parameter counts remain high. In contrast, as shown in Table \ref{tab:param}, our NeurSTT, inspired by physical information, uses a "downsampling" operation to mimic the low-rank nature of infrared background estimation. This approach uses fewer parameters, resulting in relatively faster running times than other multi-frame DL methods. Although there may be a gap when compared to lighter methods, the excellent detection performance and no-training-label requirement make the running time of our NeurSTT acceptable.
\begin{table}[t!]
\setlength{\abovecaptionskip}{0pt}
\setlength{\belowcaptionskip}{0.5pt}
    \caption{The comparison of parameter amount of 3DSTPM and proposed NeurSTT when dealing with different sizes datasets.}
    \centering
    {\scriptsize
    \begin{tabular}{p{2.2cm}<{\centering}p{2.8cm}<{\centering}p{2.8cm}<{\centering}}
    \hline
       \multirow{2}{*}{Methods}  & \multicolumn{2}{c}{Parameter Amount}\\\cline{2-3}
        & $256\times256$ & $720\times480$\\\hline
       3DSTPM  \citep{zhang20233dstpm} &  5.269 M &26.404 M\\
       \rowcolor{gray!20} NeurSTT  & 0.317 M {\textcolor{Orchid}{{\scriptsize ($16.6\times$ Fewer)}}}& 2.117 M {\textcolor{Orchid}{{\scriptsize ($12.5\times$ Fewer)}}}\\
       \hline
    \end{tabular}}
        \vspace{-0.25cm}
    \label{tab:param}
\end{table}
\begin{figure}[t!]
\setlength{\abovecaptionskip}{-7pt}
\setlength{\belowcaptionskip}{-2pt}
\centering
  \includegraphics[width=\linewidth]{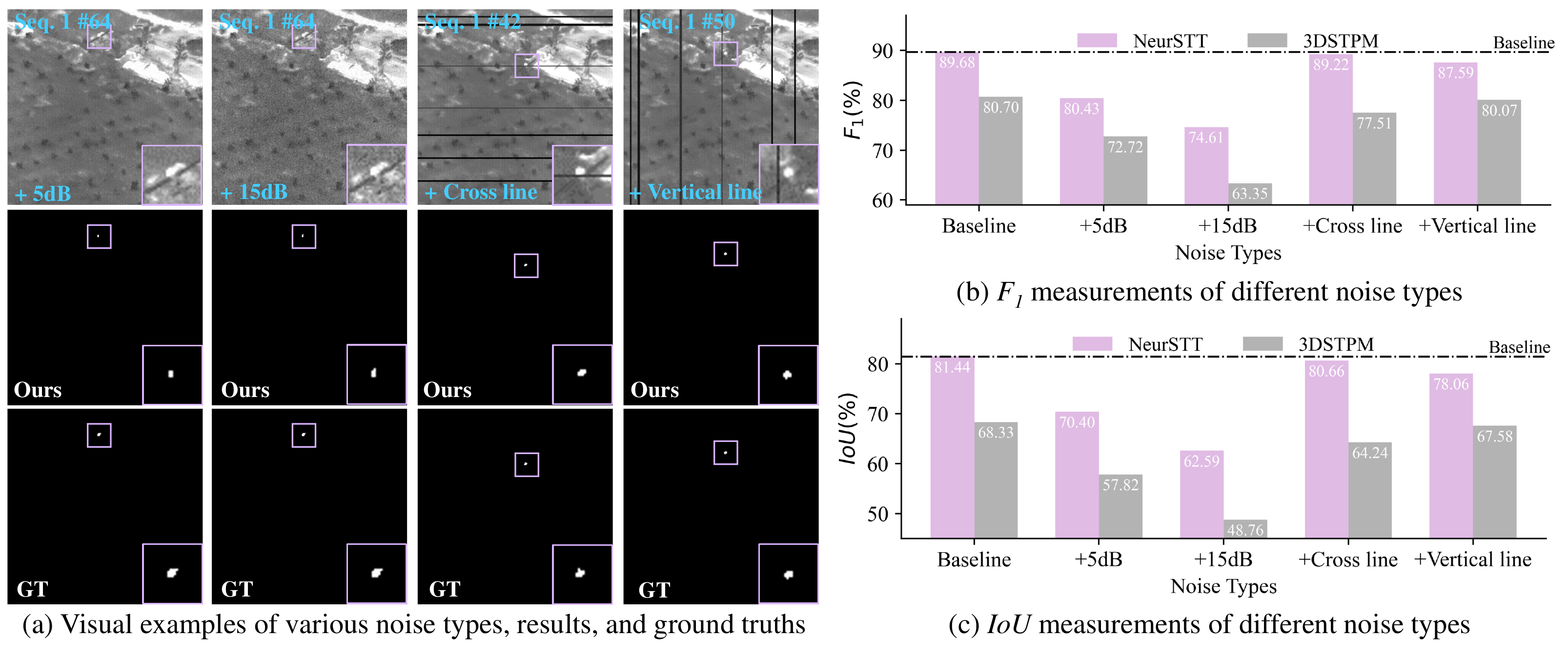}%
  \caption{(a) Visualized examples of different noise interferences on Seq. 1, along with corresponding results and ground truth labels; (b) and (c) show the corresponding averaged $F_1$ and $IoU$ evaluations on six sequences from \citep{hui2019atr}.}
    \label{fig:noise_results}
    \vspace{-0.2cm}
\end{figure}
\subsection{Robustness of Various Scenes}
In ISTD, the robustness of algorithms determines their suitability for widespread use. In this section, we test the robustness of our model by adding different types of noise. We select six sequences from \citep{hui2019atr} and introduce four types of noise: 5dB and 15dB white Gaussian noise, and randomly placed cross and vertical noises in various shapes, as presented in Fig. \ref{fig:noise_results}(a). We also present our model's numerical performance ($IoU$ and $F_1$) against various noises and include 3DSTPM for comparison in Figs. \ref{fig:noise_results}(b) and (c).

We can observe that our method maintains most of its performance when introducing cross or vertical lines. When encountering higher levels of Gaussian noise, it may experience slight performance loss. However, as shown in Fig. \ref{fig:noise_results}(b), compared to 3DSTPM, which suffers a significant decline of 28.64\% when noise increases to 15dB, our model only declines by 23.31\% in $F_1$. Thus, our model demonstrates better noise resistance and robustness.

\section{Conclusion}
\label{sec6}
In this article, we design a neural-represented spatial-temporal tensor model for ISTD using an unsupervised learning approach. Inspired by low-rank properties and the nonlinearity of neural networks, we propose a continuous low-rank neural representation module for background estimation. To leverage temporal information, we design a three-dimensional neural total variation model for regularizing backgrounds and capturing spatial-temporal features. We use the traditional soft-thresholding operator as the target solver, integrating it into the final loss function. Ablation studies and superior performance on various datasets confirm NeurSTT's effectiveness in ISTD.

In future work, we plan to optimize our model more flexibly to enhance its effectiveness and efficiency, given the adaptability of neural networks. Meanwhile, we aim to "neuralize" more advanced tensor decomposition methods to improve the correlation of spatial-temporal features, thereby better assisting in detection performance.

\section*{Acknowledgments}
We thank Y. Luo for his rigorous derivations and code implementation of mathematical models.

\appendix
\section{Deductions of Function Network's Derivative}
\label{appendix}
For a input vector $\textbf{h} \in \mathbb{R}^{n_1}$, given the neural network function $f_{\theta_h}$, we calculate its derivative as follows: 
\begin{equation}
\setlength{\abovedisplayskip}{-5pt}
\setlength{\belowdisplayskip}{-5pt}
\begin{aligned}
f_{\theta_h}(\textbf{h}) = \textbf{H}_2 \sin(\textbf{H}_1 \sin(\textbf{H}_0 \cdot \textbf{h}))
\end{aligned}    
\end{equation}
We apply the chain rule to solve it.
Initially, we let $\textbf{u} = \textbf{H}_1 \sin(\textbf{H}_0 \cdot\textbf{h})$. 
Then we have $f_{\theta_h}(\textbf{h}) = \textbf{H}_2 \sin(\textbf{u})$, where the derivative with respect to $\textbf{u}$ is:
\begin{equation}
\setlength{\abovedisplayskip}{-2pt}
\setlength{\belowdisplayskip}{-2pt}
\begin{aligned}
\frac{\partial}{\partial\textbf{u}} (\textbf{H}_2 \sin(\textbf{u})) = \textbf{H}_2 \cos(\textbf{u})
\end{aligned}    
\end{equation}
Next, we differentiate $\textbf{u}$ with respect to $\textbf{H}$: Let $\textbf{v} = \textbf{H}_0\cdot\textbf{h}$. Thus, $\textbf{u} = \textbf{H}_1 \sin(\textbf{v})$. The derivative of $\textbf{u}$ with respect to $\textbf{v}$ is: 
\begin{equation}
\setlength{\abovedisplayskip}{-2pt}
\setlength{\belowdisplayskip}{-2pt}
\begin{aligned}
\frac{\partial}{\partial\textbf{v}} (\textbf{H}_1 \sin(\textbf{v})) = \textbf{H}_1 \cos(\textbf{v})
\end{aligned}    
\end{equation}
Differentiate $\textbf{v}$ with respect to $\textbf{H}$:
\begin{equation}
\setlength{\abovedisplayskip}{-2pt}
\setlength{\belowdisplayskip}{-2pt}
\begin{aligned}
\frac{\partial}{\partial\textbf{h}} (\textbf{H}_0 \cdot\textbf{h}) = \textbf{H}_0
\end{aligned}    
\end{equation}
Combine the derivatives using the chain rule: 
\begin{equation}
\setlength{\abovedisplayskip}{-2pt}
\setlength{\belowdisplayskip}{-2pt}
\begin{aligned}
\frac{\partial f_{\theta_h}(\textbf{h})}{\partial\textbf{h}} = \frac{\partial f_{\theta_h}(\textbf{h})}{\partial\textbf{u}} \cdot \frac{\partial\textbf{u}}{\partial\textbf{v}} \cdot \frac{\partial\textbf{v}}{\partial\textbf{h}}
\end{aligned}    
\end{equation}
Substituting the derivatives we found, we can have the final derivative: 
\begin{equation}
\setlength{\abovedisplayskip}{-2pt}
\setlength{\belowdisplayskip}{-2pt}
\begin{aligned}
\nabla_h {f_\Theta(\textbf{h})} \!=\!\frac{\partial f_{\theta_h}(\textbf{h})}{\partial\textbf{h}} = \textbf{H}_2 \cos(\textbf{H}_1 \sin(\textbf{H}_0\cdot \textbf{h}))\cdot \textbf{H}_1 \cos(\textbf{H}_0 \cdot \textbf{h}) \cdot \textbf{H}_0
\end{aligned}    
\end{equation}

 \bibliographystyle{elsarticle-num} 
 \bibliography{cas-refs}




\end{document}